\documentclass[lettersize,journal]{IEEEtran}
\usepackage{amsmath,amsfonts}
\usepackage{amssymb}
\usepackage{amsthm}
\usepackage{algorithmic}
\usepackage{algorithm}
\usepackage{array}
\usepackage[caption=false,font=normalsize,labelfont=sf,textfont=sf]{subfig}
\usepackage{textcomp}
\usepackage{stfloats}
\usepackage{url}
\usepackage{verbatim}
\usepackage{graphicx}
\usepackage{cite}
\usepackage{booktabs}
\theoremstyle{plain}
\newtheorem{theorem}{Theorem}

\theoremstyle{definition}
\newtheorem{definition}[theorem]{Definition}
\newtheorem{assumption}[theorem]{Assumption}
\theoremstyle{remark}

\begin{document}

\title{MSfusion: A Dynamic Model Splitting Approach for Resource-Constrained Machines to Collaboratively Train Larger Models}

\author{Jin Xie,
        Songze Li
        % <-this % stops a space
%\thanks{Manuscript received 4 July, 2024;}
\thanks{Jin Xie is with the Internet of Things Thrust, The
Hong Kong University of Science and Technology (Guangzhou), Guangzhou, China (email: jxie171@connect.hkust-gz.edu.cn). 

Songze Li is with the School of Cyber Science and Engineering, Southeast
University, Nanjing, China (e-mail: songzeli@seu.edu.cn).}% <-this % stops a space
}

% \markboth{IEEE TRANSACTIONS ON MOBILE COMPUTING, JULY~2024}%

% \IEEEpubid{0000--0000/00\$00.00~\copyright~2024 IEEE}

\maketitle

\begin{abstract}
Training large models requires a large amount of data, as well as abundant computation resources. While collaborative learning (e.g., federated learning) provides a promising paradigm to harness collective data from many participants, training large models remains a major challenge for participants with limited resources like mobile devices. We introduce {\ttfamily MSfusion}, an effective and efficient collaborative learning framework, tailored for training larger models on resource-constraint machines through \emph{model splitting}. Specifically, a double shifting model splitting scheme is designed such that in each training round, each participant is assigned a subset of model parameters to train over local data, and aggregates with sub-models of other peers on common parameters. While model splitting significantly reduces the computation and communication costs of individual participants, additional novel designs on adaptive model overlapping and contrastive loss functions help {\ttfamily MSfusion} to maintain training effectiveness, against model shift across participants. Extensive experiments on image and NLP tasks illustrate significant advantages of {\ttfamily MSfusion} in performance and efficiency for training large models, and its strong scalability: computation cost of each participant reduces significantly as the number of participants increases.
\end{abstract}


\section{Introduction}
\IEEEPARstart{T}he contemporary technological landscape, characterized by the emergence and rapid evolution of large models, has ushered in a transformative era for machine learning and artificial intelligence. \cite{brown2020language, floridi2020gpt, zhang2022opt, chowdhery2023palm, achiam2023gpt} Large language models (LLMs), exemplified by GPT and its counterparts, trained on vast corpora of billions of tokens, have captivated the global community with their remarkable capabilities, including human-like text generation, language translation, question answering, etc \cite{radford2018improving, devlin2018bert, raffel2020exploring, bommasani2021opportunities, zhao2023survey}. However, the practical training of these models is hamstrung by substantial computational and data requirements.

Consider the following real-world scenario: multiple companies, each armed with its own resource-limited servers (or cloud instances) and the private data collected from their respective clients, aspire to harness the advantages of large foundation models. The objective, therefore, is to leverage the existing computational power of their servers collaboratively to train a high-performance large model. Additionally, due to privacy and cost considerations, the introduction of an additional central server is unsuitable for these companies. Conventional distributed learning methods like FedAVG \cite{mcmahan2017communication} is not applicable as it is not practical to perform local SGD on large models, given the memory and computation constraints on companies' local servers. 

As demonstrated in \cite{dey2023cerebrasgpt}, utilizing only $10\%$ of a large language model during training can result in up to a 100-fold reduction in Floating Point Operations (FLOPs), translating to substantial cost savings. Motivated by this, we ask the following question: \emph{Is it possible for these companies to collaboratively train a high-performance large model over their private data, with each company only training a sub-model as a split from the full model?} 

To address the above question, we propose {\ttfamily MSfusion}, a novel collaborative learning framework that utilizes model splitting to enable effective and efficient training of larger models over resource constrained participants.
{\ttfamily MSfusion} leverages a network of decentralized participants, each equipped with its own dataset, to independently extract and train split models from a larger model, effectively managing resource constraints. 
A novel double shifting splitting scheme is proposed to ensure extensive coverage of the global full model by the participants. An overlap aggregation method is introduced to further reduce communication needs. Moreover, an adaptive splitting mechanism is introduced to dynamically adjust the overlap of model parameters across participants as training progresses, expediting model convergence. A contrastive objective is designed to mitigate model drift caused by heterogeneous data distributions and differences in participants' sub-models.

\IEEEpubidadjcol

{\ttfamily MSfusion}, as a combination of model and data parallelism, not only reduces computation and communication costs, but also enhances the model performance via utilizing diverse datasets across multiple participants. We implement {\ttfamily MSfusion} and evaluate it over various image and NLP datasets. Extensive experiments demonstrate the substantial advantages of {\ttfamily MSfusion} in model performance and computation and communication efficiencies, over SOTA distributed learning methods using model splitting. 
{\ttfamily MSfusion} also exhibits strong scalability such that to achieve some target accuracy, the required split model size (hence computation/communication load) of each participant decreases significantly as the number participants increases. We view this as a key enabler for more resource-constraint participant to contribute to and benefit from training of large models.

\section{Related works}
\subsection{Decentralized Learning}

Decentralized learning, in contrast to its centralized counterpart, pursues a consensus model through peer-to-peer communication, eliminating the reliance on a central server. \cite{wang2022accelerating, yuan2024decentralized, chen2024enhancing, sun2022decentralized} Commonly used discentrailized FL like D-PSGD \cite{lian2017can} offers distinct advantages in terms of communication efficiency and data privacy preservation when compared to Centralized Federated Learning (CFL). In an exemplary serverless, peer-to-peer FL implementation, \cite{roy2019braintorrent} introduced BrainTorrent, which has found application in dynamic peer-to-peer FL environments, particularly in medical contexts. Dis-PFL is proposed in \cite{dai2022dispfl}, which employs personalized sparse masks to train personalized models, reducing communication costs by filtering out parameter weights with minimal influence on the gradient. 
%\cite{shi2023towards} have contributed to the field by enhancing representation abilities through decentralized partial model personalization, tailoring personalization in FL to better suit individual needs.

\subsection{KD-based Method}
% Knowledge Distillation (KD) based methods offer a solution wherein the complex knowledge encapsulated in a large model (server model) is imparted to a smaller, more tractable model (client model). 
In the context of minimizing computation costs for participants, Knowledge Distillation (KD) emerges as a viable option. KD methods involve the server distilling intricate knowledge from a large model into a smaller, more manageable model, enabling clients to locally train on this distilled model. \cite{gou2021knowledge, cho2019efficacy, mirzadeh2020improved, zhou2023mec}
Methods such as FedET in \cite{cho2022heterogeneous} have demonstrated some efficiency of KD. The defining strength of KD-based methods lies in their ability to train more compact models, approximating the performance of their larger counterparts with substantially less computational overhead. Nonetheless, achieving competitive accuracy typically necessitates access to public datasets that align in domain and scope with the client data \cite{lin2020ensemble}.
However, a critical constraint of KD-based methods is the necessity for a central server, rendering them less suited to the decentralized collaborative settings introduced in this study.

\subsection{Partial Model Training}
The partial model training (PT-based) paradigm presents a distributive strategy wherein the model is fragmented across multiple servers, and each server is responsible for training a discrete segment of the model \cite{hong2022efficient, alam2023fedrolex, diao2021heterofl}. This modular approach considerably alleviates the computational demands on individual servers and fosters parallelized training. A notable limitation of current PT-based methodologies is their confinement to CFL frameworks, designed to reconcile computational disparities among clients. Such methods inherently assume that some participants possess the \emph{complete} model during the training phase, a presumption misaligned with the decentralized scenario envisaged in our work. As we experimentally show later in Section \ref{Experiments}, without such participants, the performance of these methods degrade severely.
% And these participant actually play a important role, without such participants these methods failed to obtain a good performance shown in Section \ref{Experiments}.
\cite{shulgin2023towards} provide detailed analysis towards PT-based methods, further showing its potential for efficient collaborative training.

\noindent {\bf Distinction from split learning.} We emphasize that the scenario considered in this work is drastically different from that of split learning~\cite{vepakomma2018split, singh2019detailed, gao2020end, thapa2022splitfed}, another collaborative training paradigm that splits model parameters between a powerful central server and one or multiple clients. In split learning, the 
server and the clients work sequentially to train the entire model, through communicating intermediate data embeddings and gradients; a client cannot train its own sub-model independently due to missing gradients backpropagated from the server. While the key idea of split learning is to offload most of the computations on a powerful server, and requires the server and the clients to compute one after another, we focus on the scenario where all participants have similar computation capabilities, and each of them \emph{independently} trains a smaller sub-model locally.

\begin{figure*}[htbp]
\begin{center}
\centerline{\includegraphics[width=165mm]{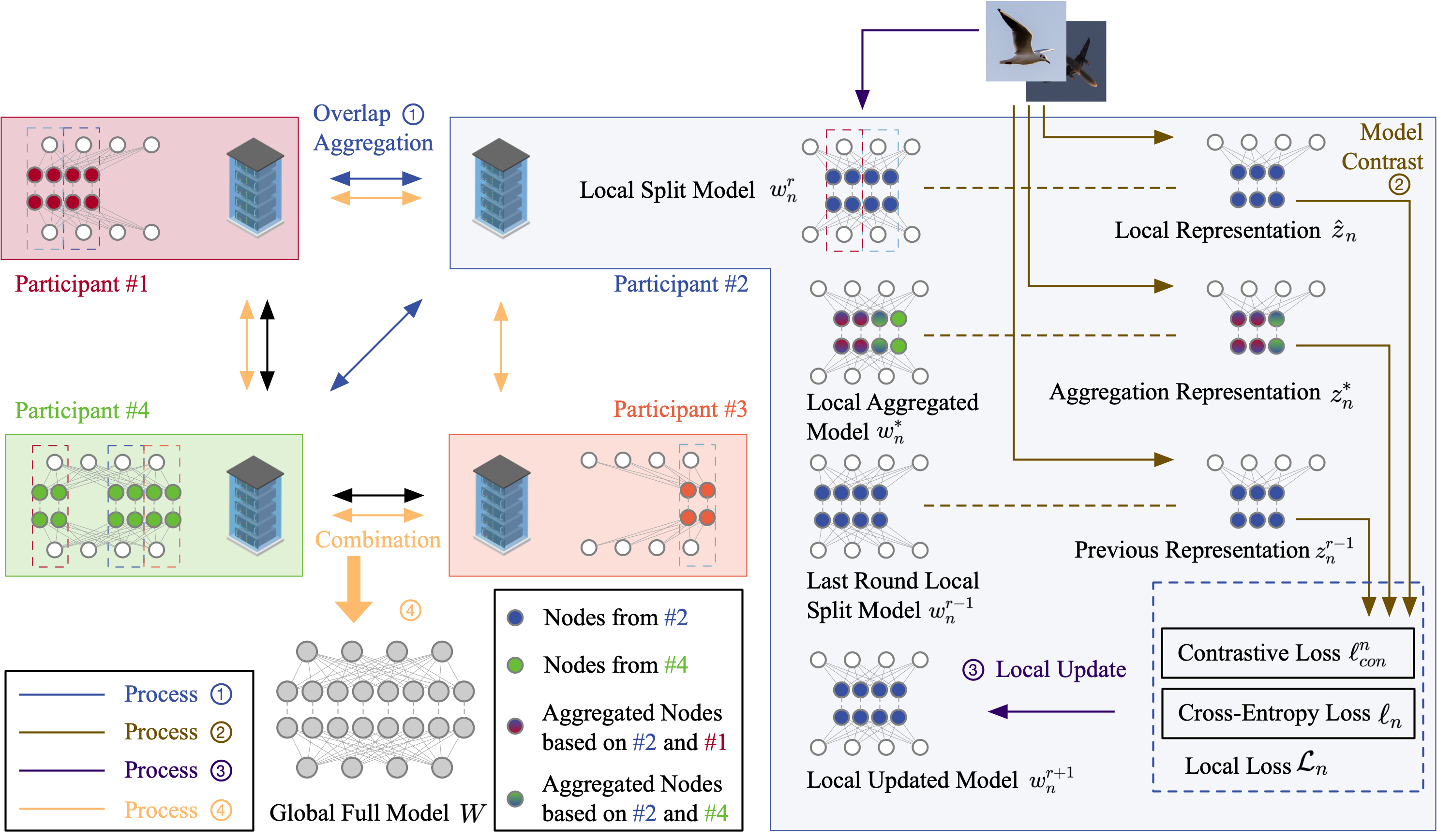}}
% \vspace{-2mm}
\caption{Overview of MSfusion.}
\vspace{-7mm}
\label{MSfusion_Dig}
\end{center}
\end{figure*}

\section{Problem Definition}
Consider a collaborative learning system of $N$ participants. Each participant $n$, $n \in [N] := \{ 1,...,N \}$, has a local dataset $\mathcal{D}_n=\{(x _i^{(n)},y_i^{(n)})\}_{i=1}^{M_n}$ with $M_n$ collected samples. The goal is to train a global model $W$ over all participants' datasets, to solve the following optimization problem:
% For a CFL training a global model $W$, its global objective is solving the following finite-sum stochastic non-convex optimization problem:
\begin{equation}
    \begin{aligned}
    \min_{W } \enspace & f(W)=\frac{1}{N} \sum_{n=1}^{N} \hat{f_n} (W),\\
\text{s.t.} \enspace &  \hat{f_n} (W) :=\frac{1}{M_n}\sum_{i=1}^{M_n}\mathcal{\hat{L} }(W;(x _i^{(n)},y_i^{(n)})).
\end{aligned}  
\label{collaborative}
\end{equation}
Here $\hat{f_n}(W)$ is the local empirical risk of participant $n$, for some loss function $\mathcal{\hat{L}}$. 

We focus on the scenarios of collaborative training of larger models on resource-constrained participants, where it is impractical for a participant to locally train $W$ due to limited memory and computation resources.
% That is we assume for each participants, the memory resources and computational power is constrained. 
We consider a model splitting framework, such that each participant trains a smaller sub-model $w_n$ split from the global full model $W$ ($w_n \subseteq W$). %based on splitting scheme toward hidden channel. 
Based on this, the local empirical risk for participant $n$ becomes
\begin{equation}
    f_n(w_n):=\frac{1}{M_n}\sum_{i=1}^{M_n}\mathcal{L}_n(w_n;(x _i^{(n)},y_i^{(n)})),
\label{DFCL_local}
\end{equation}
where ${\cal L}_n$ is the local loss corresponding to the sub-model $w_n$. After the participants finish their local training, the obtained sub-models are fused into a global model. This model fusion can take place over many rounds, and the sub-model trained at each participant can vary across rounds.
% the decentralized training process, these models are fusion with neighbors' to obtain the global full model for inference. 
% This collaboratively training solution is realistic, since in most cases, the memory consumption of storing the gradients dominates the memory consumption of storing the weights \cite{cheng2022does}.

We define \emph{split model size} of participant $n$, denoted by $\mu_n$, as the ratio of the size of $w_n$ to the size of $W$, i.e., $\mu_n = \frac{|w_n|}{|W|}$. In practice, $\mu_n $ is determined by the computation and communication capabilities of the participant. While previous studies have often assumed the presence of a powerful participant who can process the entire model, i.e., $\mu_n =1$, we focus primarily on the scenarios where a group of resource-constrained participants collaborate to train an effective large model, where all participants have comparable but small split model sizes, e.g., $\mu_n \leq 0.5$ for all $n$.  

Training large models over the collaborative learning framework described above are faced with following challenges.

\begin{itemize}
\item \textbf{Efficiency:} Computation cost for local training and communication cost to exchange models/gradients are major efficiency bottlenecks when dealing with large models \cite{chen2022fedobd, zhang2023gpt}. 
% \songze{Are these references for large models?} 
Although model splitting helps to alleviate this issue, doing it naively may significantly reduce the model performance.
\item \textbf{Data and model heterogeneity:} Like in FL, the local data on different participants tend to follow different distributions; in addition, the split portions of the global full model may diverge among participants. As it is well known that data non-iidness leads to reduced model performance~\cite{li2021ditto, fallah2020personalized, collins2021exploiting}, the ``model drift'' caused by double heterogeneity of sub-model and local data poses serious challenges on training effective large models.
\item \textbf{Scalability:} To encourage participation of more resource-limited devices, it is desirable that as the number of participants increases, a smaller split model size is required on each participant to achieve a target accuracy.  
%Scalability is also important in collaborative training, especially in real-world scenarios where an increased number of participants implies greater access to training data and the potential for improved performance. 
However, more participants exacerbates the issue of model drift, potentially degrading the model performance. How to design model splitting to maintain model performance with reduced split model size is hence crucial to achieving scalable collaborative training. 
\end{itemize}

\section{MSfusion}
\label{MSfusion}
In this section, we introduce {\ttfamily MSfusion}, a model splitting approach to address the above challenges, for effective and efficient collaborative training of large models. 
% introduce the proposed MSfusion to address the challenges involved in DFCL. 
Figure \ref{MSfusion_Dig} provides an overview of proposed {\ttfamily MSfusion}. Algorithm \ref{alg:MSfusion}
gives the pseudo-code of {\ttfamily MSfusion}.

\begin{figure*}[htbp]
\begin{center}
\centerline{\includegraphics[width=180mm]{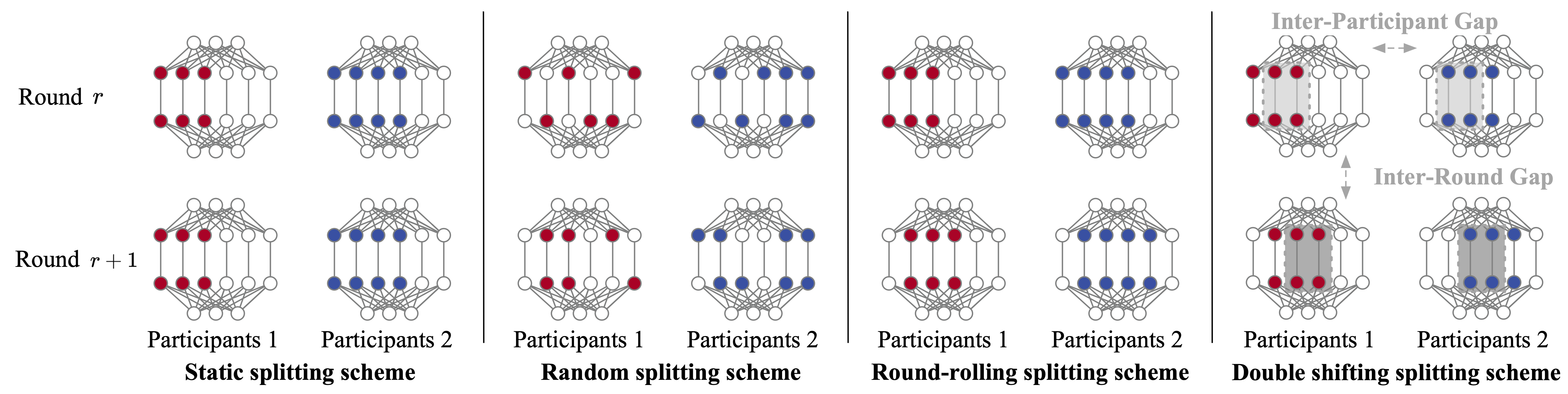}}
% \vspace{-2mm}
\caption{Difference between DSS and previous model splitting schemes.}
\label{Different_schemes}
\end{center}
\vspace{-7mm}
\end{figure*}

\subsection{Double shifting splitting scheme}
We first introduce a novel splitting approach, dubbed Double Shifting Model Splitting Scheme (DSS), as essential part of our {\ttfamily MSfusion} framework for achieving model partitioning among participants. {\ttfamily MSfusion} employs a two-tier shifting model partitioning approach: one at the inter-participant level and another at the inter-round level. This departs significantly from existing methodologies such as Federated Dropout, which utilizes random splitting; HeteroFL and FjORD, which employ static splitting; and FedRolex, which implements round-rolling splitting. %In contrast, MSfusion employs a two-tier shifting model partitioning approach: one at the inter-participants level and another at the inter-round level. 
Figure \ref{Different_schemes} shows the difference between DSS and previous model splitting schemes. 

\textbf{Inter-participant gap:} During either the initiation phase of training or when there is a change in the set of participants (either through joining or leaving), each participant is assigned an unique index. For participants with adjacent indices, the starting nodes of their sub-models at $i$-th hidden layer of the global full model differs by the following gap
    % \songze{Did we talk about network topology before?} 
% In a fully connected network, this index could be randomly assigned. For participants with nearest network topologies, the indexing is designed to be proximate, thereby facilitating optimal communication. For participants with adjacent indices, 
% \songze{for each $i$-th hidden layer of the global full model, the starting nodes of their sub-models differ by a gap}
% a designated shifting gap $G_{i}$ is introduced based on:
\begin{equation}
G_{i}=\frac{K_i}{N}\times c,
\end{equation}
where %$N_n$ denotes the total number of participants, 
$K_i$ represents the size of $i$-th hidden layer, and $c\in [0,1]$ serves as an overlapping control parameter. Consequently, the index of the starting node at participant $n$ is $nG_{i}$.
This gap ensures that the entirety of the global model $W$ is adequately represented across all participants during a single communication round. Moreover, it ensures overlap (shade part in Figure \ref{Different_schemes}) between sub-models assigned to adjacent participants. %thereby the federated overlapping average can be achieved in MSfusion. 

\textbf{Inter-round gap:} Between successive communication rounds, each participant shifts the starting node of its local sub-model by a gap $\zeta$. This gap ensures that the parameters of the global model $W$ are uniformly optimized by individual participants. 

%\textbf{DSS Formalization:} 
For a participant with index $n$ and a split model size $\mu_n$, for the $i$-th hidden layer of the global full model, the indices of the neurons (for fully-connected layers and hidden layers of attention heads of transformers) or filters (for convolutional layers) contained in the sub-model $w_n$ in round $r$, denoted by $\varphi _{n,i}^{(r)}$, are presented as follows.
% subset (split model $w_n$) of the global full model $W$ assigned to this participant can be precisely defined using our DSS formalization as:
\begin{equation}
\varphi _{n, i}^{(r)}\!\!=\!\!\left\{\begin{array}{ll}\!\!\left\{\mathcal{G}_{n,i,r} , \mathcal{G}_{n,i,r}+1, \ldots, \mathcal{G}_{n,i,r}+\left\lfloor\mu_n K_{i}\right\rfloor-1\right\},   \\
\quad\quad\quad\quad\quad\quad\quad\text { if } \mathcal{G}_{n,i,r}+\left\lfloor\mu_n K_{i}\right\rfloor \leq K_{i}, 
\\
\!\!\left\{\mathcal{G}_{n,i,r}, \mathcal{G}_{n,i,r}+1, \ldots, K_{i}-1\right\} \cup \\\left\{0,1, \ldots, \mathcal{G}_{n,i,r}+\left\lfloor\mu_n K_{i}\right\rfloor-1-K_{i}\right\}  ,\text { else. }\end{array}\right.
\label{DSS}
\end{equation}
% In this formalization, $\Theta_{n, i}^{(r)}$ represents the indices of neurons or filters in the $i$-th hidden layer of the model partition for participant $n$ at round $r$. Here, neurons pertain to fully-connected and recurrent layers, whereas filters relate to convolutional layers. 
% In the context of transformers, this refers to the neurons in the hidden layers of the attention heads. 
Here $\mathcal{G}_{n,i,r} =nG_{n,i}+r\zeta$ is the combined Inter-Participant and Inter-Round Gap. Note that in DSS the split model size $\mu $ dictates the number of output channels/neurons for the corresponding layer, which subsequently influences the number of input channels/neurons for the succeeding layer. To preserve dimensional consistency, DSS is intentionally not applied to the input and final output layers of the model. 

\subsection{Overlap aggregation}
% The introduction of {\ttfamily MSfusion} brought with it a transformative change in model splitting by leveraging the DSS. A notable innovation is the implementation of a stable, controllable overlap, expressed as $\Psi_{\{n,n+1\}, i}^{(r)} = w_{n, i}^{(r)}\cap w_{n+1, i}^{(r)}$. Such an overlap encapsulates the intersection of adjacent participant models, distinctly differentiating {\ttfamily MSfusion} from prior methodologies.

% Traditional methods, such as static and round-rolling splits, consistently retain overlaps equal to the smaller-sized participant models, so that in each communication round only this small part of global model $W$ is collaboratively aggregated. While the random splitting method yields overlaps that are unstable for participants to utilize. {\ttfamily MSfusion}, however, simplifies overlap recognition between participants and provides robust control with the overlapping control parameter, $c$. 

The introduction of DSS allows flexible designs for overlapped parameters between participants and across rounds. Specifically,
the overlap rate (i.e., the ratio between the number of overlapped parameters and the total number of parameters in the full model) can be precisely quantified as $\delta_{\{n,n+1\}} =1-\frac{c}{\mu N}$ (with $c<\mu N$) for two adjacent participants with the same $\mu$. For any pair of participants, their overlap rate $\delta$ is constrained within the range $[0, 1-\frac{c}{\mu N}]$.

%Different from FedAvg~\cite{mcmahan2017communication}, where server aggregation rely on the complete models submitted by clients. 
In each round $r$, {\ttfamily MSfusion} starts with exchanging overlapped model parameters among connected participants. That is, each participant $n$ performs the following overlap aggregation:
% every pair of connected participants only send and receive their overlapping model parameters. This is formalized as the following overlapping average:
\begin{equation}
\theta_{n, [l, i]}^{*}=\frac{1}{|S|+1} \sum_{s_i \in {S}} (\theta_{s_i, [l, i]}^{r}+\theta_{n, [l, i]}^{r}),
\label{overlapping average}
\end{equation}
where $\theta_{n, [l, i]}^{r}$ donates the $i-$th parameter of layer $l$ at participant $n$ from $w_n^r$, $S \subset [N]$ is the set of connected participants holding $\theta_{n, [l, i]}^{r}$. This tailored aggregation around overlaps considerably reduces communication overhead from aggregation on full model size, enhancing the efficiency of distributed training of large models.
% offering an efficient conduit for training LLMs in a distributed manner.

% The {\ttfamily MSfusion}'s initial setup hinges on a consensus regarding the participant index and $\mu_n$, ensuring ease in determining overlaps. 
After overlap aggregation, each participant performs local training of its assigned sub-model on private data. %During subsequent training iterations, each participant only train its split model obtained from DSS, the global full model is not stored. 

% A convergence analysis of a preliminary version of {\ttfamily MSfusion} with DSS and overlap aggregation for cross-entropy loss is provided in Appendix~\ref{Mathematical reasoning}.

% Instead, an efficient fusion mechanism within {\ttfamily MSfusion} fetches the requisite global model. Participants engage with adjacent peers to access missing parameters (complementary set $\complement_{K_i}\Theta_{n, i}^{(r)}$), and combing these with aggregated overlap parameters thus effectively obtaining the global full model $W$ for further inference.  
% \songze{Describe how model aggregation is done. Variables in (5) are not defined!}
\setlength{\textfloatsep}{5pt}
\begin{algorithm}[h]
   \caption{{\ttfamily MSfusion}}
   \label{alg:MSfusion}
\begin{algorithmic}
   \STATE {\bfseries Input:} Multi-participants set $N_n$, model split rate $\mu_n$, local dataset $\mathcal{D}_n$, initial global full model $W^0$, final stage parameter $p$.
   \STATE {\bfseries Output:} Participants maintained split model $w_n^*$, trained the global full model $W^*$.
   \STATE {\bfseries Initiation:} Assign an index to each participant based on network topology by consensus. 
   \STATE {\bfseries Per-Participant Operations:}
   \FOR {\textnormal {round} $r<R$}
   \IF {\(r \mod Q = 0\)}
   \STATE{Update $c=c_0(1-(r/R)p)$ in \(\mathcal{G}_{n,i,r}\);}
   \ENDIF
   \STATE {Split local $w_n^r$ from $W^r$ by DSS (\ref{DSS}) with $\mathcal{G}_{n,i,r}$;}
   \STATE {Transmit $\theta_{n, [l, i]}^{r}$ to all connected participants holding $i-$th parameter;}
   \STATE {Receive $\theta_{s_i, [l, i]}^{r}$ from $S$ connected participants holding $i-$th parameter;}
   \STATE {Aggregate local parameters based on (\ref{overlapping average});}
   \STATE {Update $w_n^{*}$ based on $\theta_{n, [l, i]}^{*}$, and representations $z\gets w_n^r$, $z^*\gets w_n^*$, $z^{r-1}\gets w_n^{r-1}$.}
   \STATE {Sample batch $b = \{(x _{i},y_{i})\}_{i=1}^B$ from $\mathcal{D}_n$; }
   \STATE {The combined loss: $\mathcal{L}_n= \ell_{n}+\lambda  \ell_{con}^{n}$, the contrastive loss (\ref{Con_Loss});}
   \STATE {Update: $w_n^{r+1}\gets w_n^{r}-\eta  \nabla \mathcal{L}_n$;}
   \STATE {Receive round gap model parameter $\theta_{s, [l, i]}^{r+1}$ from connected participant.}
   \ENDFOR
   \STATE {\bfseries Global model combination:}
   \STATE {Transmit $\theta_{n, [l, i]}^{R} \in w_n^R$ to connected participants lacking $i-$th parameter;}
   \STATE {Receive $\theta_{s, [l, j]}^{R} \in \complement_{W}w_n^R $ from connected participants holding $j-$th parameter;}
   \STATE {Combine $\theta_{n, [l, i]}^{R}$ with $\theta_{s, [l, j]}^{R}$ to obtain the global combined model $W^*$.}
\end{algorithmic}
\end{algorithm}

\textbf{Dynamic overlap to boost convergence.} 
%MSfusion's dynamism is further evinced in its adaptive overlap strategy, which augments convergence rates. 
As an optimization to speed up convergence, we further design a dynamic overlap strategy that periodically adjusts the overlapping control parameter $c$. For some adjustment period $Q$, {\ttfamily MSfusion} updates the parameter $c$ in every $Q$ rounds ($r\mod Q = 0$) as
\begin{equation}
c=c_0(1-(r/R)p),
\label{dy_c}
\end{equation}
where $c_0$ is the initial control parameter, $p$ determines the final overlap rate, and $R$ is the total round number. 

As training progresses, $c$ is tapered to amplify the overlap rate $\delta$, such that updating each parameter utilizes data from more participants. The main idea behind this dynamic overlap strategy is that to cover as many as global model's parameters in the early stages of the training to obtain some basic model functionalities, and then focuses on addressing the model shift problem via fine-tuning each parameter with more participants' data.  

\subsection{Mathematical reasoning}
In this section, we will provided a convergence analysis of a preliminary version of {\ttfamily MSfusion} with DSS and overlap aggregation for cross-entropy loss.

For the collaborative learning problem considered in the (\ref{collaborative}) with $N$ participants and each with its own dataset $\mathcal{D}_n=\{(x _i^{(n)},y_i^{(n)})\}_{i=1}^{M_n}\in \mathbb{R} ^d\times \mathbb{R}$ . It can be summarized as the following problem: 
\begin{equation}
    \begin{aligned}
    \min_{W } \enspace & f(W)=\frac{1}{N} \sum_{n=1}^{N} f_n(w_n),\\
\text{s.t.} \enspace &  f_n(w_n):=\frac{1}{K}\sum_{i=1}^{K}\mathcal{L}_n(w_n;(x _i,y_i)),\\
&\quad\quad w_n \subseteq W.
\end{aligned} 
\label{DFCL}
\end{equation}

In order to better investigate the DSS scheme, the definition of unbiased compressor is introduced following \cite{beznosikov2020biased, shulgin2023towards}. 

\begin{definition}
\label{compressor}
Let $\zeta \ge 1$ and $\forall x\in \mathbb{R}^d$, for a (possibly random) mapping $\mathcal{C} :\mathbb{R}^d\to \mathbb{R}^d$, we say $\mathcal{C}$ is unbiased compressor operators ($\mathcal{C} \in \mathbb{U} (\zeta )$) if the following holds: 
\begin{equation}
    \mathbb{E}[\mathcal{C}(x)]=x, \quad \mathbb{E}\left[\|\mathcal{C}(x)\|^{2}\right] \leq \zeta \|x\|^{2}.
\end{equation}
\end{definition}

Like for a random splitting scheme splitting $q\in [d]:=\{1,...,d\}$ splitting the full model $W$, it can be viewed as a operator achieving the following
\begin{equation}
   \mathcal{C}_{\text {Random }}(W):=\mathbf{C}_{\text {Rand}} W=\frac{d}{q} \sum_{i \in Q} e_{i} e_{i}^{\top} W,
\end{equation}
where in random splitting scheme $Q\subseteq [d]$ is $q$ random sampling (a subset of $[d]$ of cardininality $q$ random chosen uniformly), $e_{1},...,e_{d}$ are standard unit basis vectors, and $\mathcal{C}_{\text {Random }}(W)$ belongs to $\mathbb{U} (\frac{d}{q} )$, for a smaller size of split model (lower $q$), the higher the variance $\zeta$ of the compressor.

The Stochastic Gradient Descent (SGD) for participant $n$ with local model $w_n^r$ in round $r$ can be written as:
    \begin{equation}
        w_n^{r+1} :=  w_n^{r}-\eta \nabla \mathcal{L}_n(w_n^{r}),
        \label{SGD}
    \end{equation}
where $\eta$ is the step size. In this paper, we consider the splitting scheme offers a sketch compressor $\mathbf{C}_n^r \in \mathbb{R}^d\times \mathbb{R}$ to achieve sketching on global full model $W$. And the split submodel computation can be represented as the following:
    \begin{equation}
        W^{r+1}=\mathbf{C}_n^rW^{r}-\eta\mathbf{C}_n^r \nabla \mathcal{L}_n(\mathbf{C}_n^rW^{r}).
    \label{split SGD}
    \end{equation}
The sketch $\mathbf{C}_n^r$ requires to be symmetric positive semi-definite matrix. The ideal of (\ref{split SGD}) is to reduce the cost of directly computation on full model $W^{r}$ since the compressor $\mathbf{C}_n^r$ ensure the update lies in a lower dimensional subspace. Each participant only compute its own the split model $w_n^{r}=\mathbf{C}_n^rW^{r}$, and we need this split model can effectively representing the global full model.

We mainly discuss the case where the global full model size is smaller than split model times participants number ($1 \le  \mu N$). In this situation, the Inter-Participant gap in DSS scheme ensures that the global full model $W$ can be fully covered by participants in each round. That is in round $r$, we have inter-participant compressor
    \begin{equation}
        \mathbf{C}_{\text{part}}^{r} :=\frac{1}{n} \sum_{i=1}^{n} \mathbf{C}_i^{r}=\mathbf{I} ,
        \label{round}
    \end{equation}
where $\mathbf{I} $ is the identity matrix. And this means that in each round viewed from all the participants, the (\ref{split SGD}) is equivalent to the (\ref{SGD}). 

For the inter-round gap in DSS scheme ensures that the parameters of the global model $W$ are uniformly optimized by individual participants. That is for participant $n$, there exist $r^*$ to let the inter-round compressor
    \begin{equation}
        \mathbf{C}_{\text{round}}^{n} :=\frac{1}{q^*} \sum_{i=1}^{r^*} \mathbf{C}_n^{i}=\mathbf{I} ,
        \label{round_gap}
    \end{equation}
where $q^*\in [d]$ is determined by the split model rate $\mu$. And this means means that for each participants viewed from finite round $r^*$, the (\ref{split SGD}) is equivalent to the (\ref{SGD}).  

Since from the Definition. \ref{compressor}, $\mathbb{E}\left[\|\mathcal{C}(x)\|^{2}\right] \leq \zeta \|x\|^{2}$, the overlapping parts can be viewed as an other unbiased overlapping compressor $\mathbf{C}_{\text{over}}^{n}$:

    \begin{equation}
        \mathbf{C}_{\text{over}}^{n} :=N\cdot \sum_{j\in S}e_{\pi_j}  e_{\pi_j} ^{\top},
        \label{overlapping}
    \end{equation}
where $\pi_j$ is subset of $[d]$ determined by the overlapping part between participant $j$ and participant $n$, and in {\ttfamily MSfusion} this can be controled by $c$ in $dy_c$. 
$S\subseteq \in N$ is a set of participants connected with $n$ and have overlapping part. 

Then we will make some commonly used assumptions to facilitate the analysis.

% For the case where model size is greater than split model times participants number ($1>\mu N$), in ${\ttfamily MSfusion}$ overlapping compressor (\ref{overlapping}) is also ensured, (\ref{round_gap}) still holds, and the Inter-Participant compressor

\begin{assumption}
The local loss function $\mathcal{L}_n$ is $\mathbf{L}$-smooth (matrix smoothness \cite{safaryan2021smoothness, wang2022theoretically}) and $u$ strongly convex. That is $\forall x, y \in \mathbb{R}^{d}$, there exist a positive semi-definite matrix 
$ \mathbf{L}\succeq 0$
    \begin{equation}
\mathcal{L}_n(y) \leq \mathcal{L}_n(x)+\langle\nabla \mathcal{L}_n(x), y-x\rangle+\frac{1}{2}\langle \mathbf{L}(y-x),y-x  \rangle ,
        \label{smooth}
    \end{equation}
 and for some $u>0$
    \begin{equation}
\mathcal{L}_n(y) \ge  \mathcal{L}_n(x)+\langle\nabla \mathcal{L}_n(x), y-x\rangle+\frac{u}{2}\|y-x\|_{2}^{2} .
        \label{strongly}
    \end{equation}
\end{assumption}

The standard (scalar) smoothness ($L$-smoothness) can be viewed as a special case of (\ref{smooth}) with $\mathbf{L} =L\cdot \mathbf{I} $.

\begin{assumption}
In each communication round $r$, all the participant can computes the true gradient $\mathbf{C}_n^r \nabla \mathcal{L}_n(\mathbf{C}_n^rW^{r})$ through its local submodel $w_n^{r}=\mathbf{C}_n^rW^{r}$.
\end{assumption}
To better study properties for DSS, we simplify the problem (\ref{DFCL}) into a quadratic problem 
     \begin{equation}
f(W)=\frac{1}{N} \sum_{n=1}^{N} f_{i}(w_n), \quad f_{n}(w_n) = \frac{1}{2} w_n^{\top} \mathbf{L}_{n} w_n-w_n^{\top} \mathrm{b}_{i}.
        \label{quadratic}
    \end{equation}
And under this simplification, $f(x)$ is $\overline{\mathbf{L}} $-smooth, and $\nabla f=\overline{\mathbf{L}}x-\overline{b}$ with $\overline{\mathbf{L}} =\frac{1}{n} \sum_{n=1}^{N} \mathbf{L}_i$ and $\overline{b}=\frac{1}{n} \sum_{n=1}^{N} b_i$.

We mainly examine the case of $b_i\equiv 0$, in this situation, the overall updating can be written as:
     \begin{equation}
\frac{1}{N} \sum_{n=1}^{N} \mathbf{C}_n^r \nabla f_{i}(\mathbf{C}_n^rw_n)=\frac{1}{N} \sum_{n=1}^{N} \mathbf{C}_n^r \mathbf{L}_i\mathbf{C}_n^rw_n=\overline{\mathbf{B}} ^rw_n^r .
        \label{com_grad}
    \end{equation}
Proved in \cite{shulgin2023towards}, we have the following theorem.
\begin{theorem}
\label{convergence}
\cite{shulgin2023towards}
Consider a distributed learning setting with learning process shown in (\ref{com_grad}) for a quadratic problem (\ref{quadratic}) with $\overline{\mathbf{L}} \succ 0 $ and $b_i\equiv 0$. Then for 
$\overline{A} :=\frac{1}{2} \mathbb{E} [\overline{\mathbf{L}}\overline{\mathbf{B}}^r+\overline{\mathbf{L}}^r\overline{\mathbf{B}}]\succ 0$, there exists a constant $\xi > 0$. 
     \begin{equation}
\mathbb{E} [\overline{\mathbf{B}}^r\overline{\mathbf{L}}\overline{\mathbf{B}}^r]\preceq \xi\overline{A},
        \label{EBLB}
    \end{equation}
and for a step size $\eta (0<\eta<\frac{1}{\xi} )$ the iterates satisfy the following:
     \begin{equation}
\frac{1}{R} \sum_{r=0}^{R-1} \mathbb{E}\left[\left\|\nabla f\left(w_n^{r}\right)\right\|_{\overline{\mathbf{L}}^{-1} \overline{\mathbf{A}} \overline{\mathbf{L}}^{-1}}^{2}\right] \leq \frac{2\left(f\left(w_n^{0}\right)-\mathbb{E}\left[f\left(w_n^{R}\right)\right]\right)}{\eta R}  , 
\label{conver_1}
    \end{equation}

and
     \begin{equation}
\begin{aligned}
\mathbb{E}&\left[\left\|w_n^{r}- w_n^{*}\right\|_{\overline{\mathbf{L}}}^{2}  \right] \leq \\
&\left(1-\eta \lambda_{\min }\left(\overline{\mathbf{L}}^{-\frac{1}{2}} \overline{\mathbf{A}} \overline{\mathbf{L}}^{-\frac{1}{2}}\right)\right)^{k}\left\|w_n^{r}-w_n^{*}\right\|_{\overline{\mathbf{L}}}^{2},
\end{aligned}
\label{conver_2}
    \end{equation}
where $\lambda_{\min }()$ denotes minimum eigenvalue, $w_n^{*}:=\arg \min f(w_n)$.

\end{theorem}
By applying Theorem \ref{convergence}, inter-participant compressor where $\mathbf{C}_{\text{part}}^{r}  =\mathbf{I}$, we have $\overline{\mathbf{B}} ^r=\overline{\mathbf{L}}$, $\overline{\mathbf{B}}^r\overline{\mathbf{L}}\overline{\mathbf{B}} ^r=\overline{\mathbf{L}}^3$ and $\overline{\mathbf{A}}=\overline{\mathbf{L}}^2\succ 0$. So the (\ref{EBLB}) is satisfied for constant $\xi = \lambda_{max}(\overline{\mathbf{L}})$. And with a step size $\eta=\frac{1}{\xi} $, from (\ref{conver_1}) we have 
     \begin{equation}
\frac{1}{R} \sum_{r=0}^{R-1} \left \| \nabla f(w_n^r) \right \| _\mathbf{I}^2 \leq \frac{2\lambda_{max}(\overline{\mathbf{L}})(f\left(w_n^{0}\right)-f\left(w_n^{R}\right))}{R}   .
\label{inter-participant conver}
    \end{equation}
Which means with inter-participant compressor viewed from all the participants the problem will converge. And the analysis is the same for inter-round compressor.

Then is the analysis for overlapping compressor $\mathbf{C}_{\text{over}}^{n}$. For the case each participant with same split model size $\mu$, we have $f_{n}(w_n^r) = \frac{1}{2} {w_n^r}^{\top} \mathbf{L} w_n^r$ with $\mathbf{L} \equiv  \mathbf{L}_{n}$. If we define a diagonal matrix $\mathbf{D} =\mathrm{diag}( \mathbf{L} )$. And then \ref{quadratic} can be changed into 
     \begin{equation}
\begin{aligned} 
f_n(\mathbf{D}^{-\frac{1}{2} }w_n^r)=&\frac{1}{2}(\mathbf{D}^{-\frac{1}{2} }w_n^r)^{\top}\mathbf{L}(\mathbf{D}^{-\frac{1}{2} }w_n^r)\\=
&\frac{1}{2}{(w_n^r)}^{\top}(\mathbf{D}^{-\frac{1}{2} }\mathbf{L}\mathbf{D}^{-\frac{1}{2} })w_n^r=\frac{1}{2}{(w_n^r)}^{\top}\hat{\mathbf{L}}w_n^r  ,  
\end{aligned}
\label{newqu}
    \end{equation}
where  $\hat{\mathbf{L}}\succ0$ as  ${\mathbf{L}}\succ0$, and  $\mathrm{diag} (\hat{\mathbf{L}})=\mathbf{I} $.
Since for each participant the overlapping rate with other participant is the same. Here we mainly analysis the overlapping part between each two neighbor participant with biggest overlapping rate, then for overlapping compressor at each round $r$ we have $\mathbf{C}_{n}^{r} =N\cdot e_{\pi_n^r}  e_{\pi_n^r} ^{\top}$, where $\pi_n^r$ is the overlapping part between $n$ and its neighbor with biggest overlapping rate.  So we have 
     \begin{equation}
\mathbb{E}\left [ \overline{\mathbf{B}}^r \right ] =\mathbb{E}\left [ \frac{1}{N} \sum_{n=1}^{N} \mathbf{C}_n^r \hat{\mathbf{L}}_i\mathbf{C}_n^r \right ] =N\cdot \mathrm{diag}(\hat{\mathbf{L}})=N\mathbf{I}    .
\label{conv3}
    \end{equation}
Then the \ref{EBLB} can be transformed as 
     \begin{equation}
\xi\hat{\mathbf{A}}=\xi N\mathbf{I} \succeq N^2\mathbf{I}   ,
\label{conv4}
    \end{equation}
which holds with $\xi\ge N$. And the \ref{conver_1} can be converted to
    \begin{equation}
\begin{aligned} 
&\left\|\nabla f\left(w_n^r\right)\right\|_{\hat{\mathbf{L}}^{-1} \hat{\mathbf{A}} \hat{\mathbf{L}}^{-1}}^{2} \geq \\
 &N \lambda_{\min }\left(\hat{\mathbf{L}}^{-1}\right)\left\|\nabla f\left(w_n^r\right)\right\|_{\mathbf{I}}^{2}=N \lambda_{\max }(\hat{\mathbf{L}})\left\|\nabla f\left(w_n^r\right)\right\|_{\mathbf{I}}^{2}  . 
\end{aligned}
\label{conv5}
    \end{equation}

So the convergence guarantee for the overlapping compressor is
     \begin{equation}
\frac{1}{R} \sum_{r=0}^{R-1} \left \| \nabla f(w_n^r) \right \| _\mathbf{I}^2 \leq \frac{2\lambda_{max}(\hat{\mathbf{L}})(f\left(w_n^{0}\right)-\mathbb{E} \left [ f\left(w_n^{R}\right) \right ] )}{R}       .
\label{overlapping conver}
    \end{equation}

\subsection{Contrastive objective}
In the considered collaborative learning scenario, data on different participants often have distinct distributions, leaning to model shift after local training. In addition,   
% Heterogeneous data distributions incites a model drift among participants. In the context of the collaborative learning problem, 
model splitting across participants 
further exacerbate this drift, potentially undermining model performance. Other than dynamically adjusting the model overlap, we propose to adopt contrastive learning techniques~\cite{chen2020simple, he2020momentum, khosla2020supervised} to further mitigate the model drift problem.
% In the {\ttfamily MSfusion} framework, the local aggregated model at each participant can be viewed as a potent surrogate for the global full model. This feature inherently lends itself to the adoption of a contrastive learning strategy \cite{chen2020simple, he2020momentum}. 
However a direct application of contrastive FL methods from systems like MOON \cite{li2021model} or CreamFL \cite{yu2023multimodal} is untenable, since it is computationally infeasible to run global-local model contrast at the scale of full model size.
% since run the global-local full model contrast require much more computation power and given the absence of a centralized server orchestrating a global model in {\ttfamily MSfusion}. 

In {\ttfamily MSfusion}, we apply contrastive learning on sub-models to curb the divergence 
% the contrastive objective serves a pivotal role in curbing the divergence 
between a participant's local model and the corresponding aggregated model. For any input $x$, {\ttfamily MSfusion} extracts its local representation $\hat{z} _n$ from the current sub-model $w_n^r$, the aggregation representation $z_n^{*}$ from the local aggregated model $w_n^{*}$, and the previous representation $z_n^{r-1}$ from the sub-model of last round $w_n^{r-1}$. Note there exist a inter-round shift between sub-models in consecutive rounds, and we focus on representations on the common part between $w_n^r$ and $w_n^{r-1}$. 
% These representations is the comment part between $w_n^r$ and $w_n^{r-1}$. Given $\zeta=1$ in {\ttfamily MSfusion}, this commonality is extensive, encompassing almost the entirety of the split models. 
We construct the contrastive loss in {\ttfamily MSfusion} as follows:
\begin{equation}
\begin{aligned}
&\ell_{c o n}^{n}= \\
&-\log \frac{\exp \{[(\hat{z}_n)^T\cdot  z_n^{*}] / \tau\}}{ \exp \{[(\hat{z}_n)^T\cdot  z_n^{*}] / \tau\}+\exp \{[(\hat{z}_n)^T\cdot  z_n^{r-1}] / \tau\}},
\end{aligned}
\label{Con_Loss}
\end{equation}
where $\tau$ is the temperature coefficient to control the penalties on hard negative pairs \cite{wang2021understanding}. This loss metric is optimized to align the local sub-model's representation closely with that of the aggregated model, thereby curtailing model drift. 

The overall objective for participant $n$ is then given by:
\begin{equation}
\mathcal{L}_n= \ell(w_n^r;x,y)+\lambda  \ell_{con}^{n}.
\label{loss_MSfusion}
\end{equation}
Here $\ell$ is cross-entropy loss, and $\lambda$ is the coefficient governing the weight of the contrastive loss. 

\begin{table*}[h]
% \vspace{-1mm}
\caption{Details about datasets and experiment configurations.}
% \vspace{-3mm}
\begin{center}
\scalebox{0.87}{\begin{tabular}{lcccccc}
\toprule
\multicolumn{1}{c}{}                                                                              & CIFAR100                & TinyImageNet            & PennTreebank             & WikiText2                & WikiText103       & WikiText103 (1B model)       \\
\midrule
Data/Token size                                                                                &50,000                 & 100,000                 & 929,500                  & 2,088,600                & 103,227,000           & 103,227,000   \\
\begin{tabular}[c]{@{}l@{}}Local data/token size\\ (10 participants)\end{tabular}                    & 5,000                   & 10,000                  & 92,950                   & 208,860                  & 10,322,700         & 10,322,700      \\
Local epoch                                                                       & 1                       & 1                       & 1                       & 1                        & 1                        & 1                        \\
Batch size                                                                                             & 10                      & 40                      & 100                      & 100                      & 300  & 300                    \\
Model applied                                                                                     & ResNet18                & ResNet18                & Transformer              & Transformer              & Transformer & Transformer              \\
Hidden size                                                                        & {[}64, 128, 256, 512{]} & {[}64, 128, 256, 512{]} & {[}512, 512, 512, 512{]}  & {[}512, 512, 512, 512{]} & {[}1024, 1024, 1024, 1024{]}& {[}512, 512, 512, 512{]}\\
Embedding Size                                                                    & \multicolumn{2}{c}{N/A}                                                     & 256                      & 256                      & 1024              & 1792        \\
Number of heads                                                                   & \multicolumn{2}{c}{N/A}                                                     & 8                        & 8                        & 16 & 8                        \\
Dropout                                                                           & \multicolumn{2}{c}{N/A}                                                     & 0.2                      & 0.2                      & 0.2         & 0.2             \\
Sequence length                                                                   & \multicolumn{2}{c}{N/A}                                                     & 64         & 64              & 64                       & 64                       \\
\begin{tabular}[c]{@{}l@{}}Parameter size of \\ global full model\end{tabular}                       & 11.2M                   & 11.3M                   & 7.32M                    & 19.3M                    & 574.92M &  1.021B \\ 
\bottomrule
\end{tabular}%
}
\label{Experiments details}
\end{center}
%\vskip -0.15in
\end{table*}

\begin{table*}[h]

\caption{Global model accuracy and computation cost comparison. Since Fed-ET, FedHM and DepthFL does not offer directly implement for language modeling tasks, them are not compared in NLP tasks. FLOPs denotes the floating operations for each participant per round. Size denotes the local split model size for each participant. Experiments are performed on 10 participants. %with 100\% selected rate.}
}
% \vspace{-1mm}
\begin{center}
\scalebox{0.87}{\begin{tabular}{lcccccccccccc}
\toprule           
           & \multicolumn{6}{c}{CIFAR100}                                                                                                    & \multicolumn{6}{c}{TinyImageNet}                                                                                         \\ \midrule
Methods    & iid                      & \multicolumn{2}{c}{non-iid}                  & FLOPs                   & \multicolumn{2}{c}{$\mu_n$} & iid                      & \multicolumn{2}{c}{non-iid}                  & FLOPs           & \multicolumn{2}{c}{$\mu_n$}  \\
HeteroFL   & 19.49~$\pm$~0.9          & \multicolumn{2}{c}{12.11 $\pm$~0.7}          & 35.76M                  & \multicolumn{2}{c}{25\%}    & 11.08~$\pm$~0.6          & \multicolumn{2}{c}{14.54~$\pm$~0.4}          & 1.75B           & \multicolumn{2}{c}{62.5\%}   \\
FedRolex   & 42.33~$\pm$~0.8          & \multicolumn{2}{c}{36.61 $\pm$~1.0}          & 35.76M                  & \multicolumn{2}{c}{25\%}    & 37.05~$\pm$~1.1          & \multicolumn{2}{c}{20.63~$\pm$~0.8}          & 1.75B           & \multicolumn{2}{c}{62.5\%}   \\
DepthFL    & 41.72 $\pm$~0.9          & \multicolumn{2}{c}{36.52 $\pm$~1.0}          & 167M                  & \multicolumn{2}{c}{30\%}    & 34.21 $\pm$~1.2          & \multicolumn{2}{c}{22.15 $\pm$~0.8}          & 1.47B           & \multicolumn{2}{c}{55\%}   \\
Fed-ET     & 41.61 $\pm$ 0.4           & \multicolumn{2}{c}{35.78~$\pm$~0.5}          & 1.09B                   & \multicolumn{2}{c}{N/A}     & 29.61~$\pm$~0.4          & \multicolumn{2}{c}{19.78~$\pm$~0.6}          & 6.12B           & \multicolumn{2}{c}{N/A}      \\
FedHM      & 47.45 $\pm$ 1.2          & \multicolumn{2}{c}{40.72~$\pm$~1.1}          & 145.3M                  & \multicolumn{2}{c}{40\%}     & 43.05~$\pm$~1.2          & \multicolumn{2}{c}{21.58~$\pm$~0.6}          & 1.43B           & \multicolumn{2}{c}{40\%}      \\
MSfusion S & 43.77 $\pm$~0.5          & \multicolumn{2}{c}{37.01~$\pm$~0.7}          & \textbf{6.95M}          & \multicolumn{2}{c}{10\%}    & 12.62~$\pm$~0.7          & \multicolumn{2}{c}{11.45~$\pm$~0.5}          & \textbf{79.79M} & \multicolumn{2}{c}{12.5\%}   \\
MSfusion M & 50.04 $\pm$~0.5          & \multicolumn{2}{c}{44.11~$\pm$~0.4}          & 22.33M                  & \multicolumn{2}{c}{18.75\%} & 39.61~$\pm$~0.6          & \multicolumn{2}{c}{20.91~$\pm$~0.6}          & 297.0M          & \multicolumn{2}{c}{25\%}     \\
MSfusion L & \textbf{60.63 $\pm$~0.4} & \multicolumn{2}{c}{\textbf{47.21~$\pm$~0.5}} & 151.7M                  & \multicolumn{2}{c}{50\%}    & \textbf{51.41~$\pm$~0.}4 & \multicolumn{2}{c}{\textbf{24.67~$\pm$~0.3}} & 1.26B           & \multicolumn{2}{c}{50\%}     \\ \midrule
           & \multicolumn{3}{c}{PennTreebank}                                        & \multicolumn{3}{c}{WikiText2}                         & \multicolumn{3}{c}{WikiText103}                                         & \multicolumn{3}{c}{WikiText103 (1B model)}     \\ \midrule
Methods    & Perplexity               & Size           & $\mu_n$                     & Perplexity              & Size            & $\mu_n$   & Perplexity               & Size             & $\mu_n$                   & Perplexity      & Size    & $\mu_n$            \\
HeteroFL   & 55.97 $\pm$ 5.4          & 5.09M          & 75\%                        & 579.05 $\pm$ 8.4        & 104.02M         & 75\%      & 763.44 $\pm$ 26           & 426.34M          & 75\%                      & \multicolumn{3}{c}{N/A}                        \\
FedRolex   & 61.52 $\pm$ 6.8          & 5.09M          & 75\%                        & 547.32 $\pm$ 45         & 104.02M         & 75\%      & 653.76~$\pm$ 35          & 426.34M          & 75\%                      & \multicolumn{3}{c}{N/A}                        \\
MSfusion S & 9.09 $\pm$ 0.7           & \textbf{1.24M} & 21.875\%                    & 44.33 $\pm$ 3.6         & \textbf{30.33M} & 18.75\%   & 21.45 $\pm$~2.8           & \textbf{121.49M} & 21.875\%                  & 6.34 $\pm$ 0.4  & \textbf{213.11M} & 21.875\%           \\
MSfusion M & 8.02 $\pm$ 0.5           & 1.43M          & 25\%                        & 5.28 $\pm$ 0.4          & 34.65M          & 25\%      & 7.94 $\pm$ 0.6           & 139.02M          & 25\%                      & 5.59 $\pm$ 0.3~ & 244.07M & 25\%               \\
MSfusion L & \textbf{3.11 $\pm$ 0.2}  & 3.13M          & 50\%                        & \textbf{3.59 $\pm$ 0.2} & 69.23M          & 50\%      & \textbf{5.24~$\pm$ 0.3}  & 281.04M          & 50\%                      & \textbf{5.03 $\pm$ 0.3}  & 495.63M & 50\%          \\
\bottomrule
\end{tabular}
}
\label{Performance_Re}
%\vspace{-3mm}
\end{center}
%\vskip -0.15in
\end{table*}

After the training process, an efficient fusion mechanism within {\ttfamily MSfusion} fetches the requisite global model. Participants engage with adjacent peers to retrieve missing parameters (complementary set $\complement_{K_i}\Theta_{n, i}^{(r)}$), and combine them with aggregated overlap parameters, obtaining the global full model $W^*$ for further inference.

\begin{figure*}[h]
% \vspace{-5mm}
\begin{center}
\centerline{\includegraphics[width=180mm]{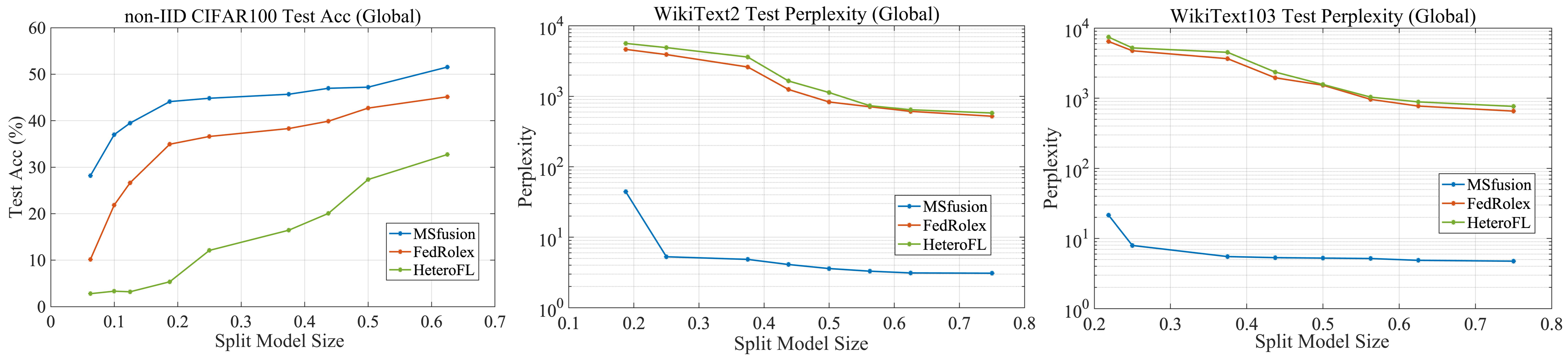}}
% \vspace{-3mm}
\caption{Performance comparisons on non-IID CIFAR100, WikiText2 and WikiText103 datasets.}
\vspace{-7mm}
\label{Performance Fig}
\end{center}
\end{figure*}

\section{Experiments}
\label{Experiments}
In this section, we present a comprehensive evaluation of {\ttfamily MSfusion} on various benchmark datasets, including both image classification and natural language processing (NLP) tasks. 
% We provide more experiment details (parameter size, data partition, and model architecture) and additional results in Appendix \ref{Experiment details and more results}.

\textbf{Datasets.} 
We conduct our evaluations on two distinct task categories: image classification and NLP. For image classification, our method is subjected to rigorous testing on two widely recognized datasets: CIFAR100 \cite{krizhevsky2009learning} and TinyImageNet\cite{le2015tiny}. 
For NLP tasks, we adopt the method in ~\cite{diao2021heterofl, alam2023fedrolex, kim2022depthfl}, dividing the data samples from the original dataset onto each participant uniformly at random. As a result, the number of possible words in each participant's local dataset is 
approximately 1,000 for PennTreebank\cite{marcus-etal-1993-building}; 3,000 for WikiText2; and 50,000 for WikiText103. However, all participants use the same table for the tokenizer, with the full vocabulary of the entire dataset.

\textbf{Models.} To underscore the versatility of our approach across varied architectures, we employ both convolutional and transformer models in our experiments. In image classification tasks, a modified ResNet18 following \cite{diao2021heterofl} is employed. Additionally, we showcase the transformative potential of {\ttfamily MSfusion} by demonstrating its efficacy in handling LLMs. Specifically, transformer models are deployed for our NLP tasks. The global full model parameter sizes for adopted transformers are 7.32M for PennTreebank, 19.3M for WikiText2, 574.92M for WikiText103, and 1.021B for WikiText103. 

\textbf{Data heterogeneity.} In the image classification tasks, we introduce data heterogeneity by deliberately skewing the label distribution among participants. This non-iid characteristic is attained by assigning each participant a distinct subset of $H$ classes. Specifically, for CIFAR100, we set $H=20$, and for TinyImageNet, $H=40$. Moreover, for NLP tasks, we naturally generate non-IID data distribution through dataset partitioning among participants. As a result, the vocabulary size is reduced for each participants. Specifically, in WikiText2, the vocabulary size is reduced from a total of 33,728 words to approximately 3,000 words. In PennTreebank, the vocabulary is reduced from 10,000 to approximately 1,000 words, and in WikiText103, it is scaled down from 267,735 to approximately 50,000 words. 

\textbf{Model splitting.} %Throughout our study, 
We focus on participants with \emph{uniformly} small model splits $\mu_n \in \{6.25\%, 10\%,..., 62.5\%\}$. The global full model represents an unsplit, complete model. \emph{There is no participant with $\mu_n=1$.} This is a notable advantage compared to previous works with heterogeneous settings. While a participant training the full model plays an important role in maintaining model performance, it is not present in our case due to large model size and the focus on resource-constrained settings.
% as it is typically unfeasible for a single participant to possess the computational resources to train the global full model and such participant plays a important role in their experiments.
To create local sub-models, we adjust the number of kernels in convolution layers of ResNet18 while keeping the output layer nodes constant. For Transformer models, we vary the number of nodes in the hidden layers of the attention heads. We set the inter-round gap for DSS $\zeta=1$, the initial overlapping control parameter $c_0=1$, and the adjustment period $Q=10$.

\textbf{Baselines.} 
We compare {\ttfamily MSfusion} against FedHM~\cite{yao2022fedhm}, DepthFL~\cite{kim2022depthfl}, SOTA PT-based model-heterogeneous FL methods including HeteroFL~\cite{diao2021heterofl} and FedRolex~\cite{alam2023fedrolex}, as well as SOTA KD-based FL method Fed-ET~\cite{cho2022heterogeneous}. %To ensure equitable comparisons, we maintain uniformity in parameters across all PT-based baselines, including learning rate, and the number of communication rounds. 
For FedHM, it leverages low-rank decomposition to reduce communication costs and training overhead. However, its reliance on server-side matrix decomposition introduces additional computational demands, particularly with large transformer models. For DepthFL, it adopts a strategy where clients train local models at depths commensurate with their resource capabilities, utilizing self-distillation to enhance the training of deeper layers.
For experiments with a ring topology, we compare {\ttfamily MSfusion} with a widely used decentralized learning algorithm D-PSGD \cite{lian2017can}, and the SOTA discentralized personalizd FL algorithm Dis-PFL \cite{dai2022dispfl} with sparse training technique. %\textcolor{red}{What is this?}

\textbf{Hyperparameters and Platform.} 
For all the experiments, SGD optimizer is applied. The communication round for CIFAR100 experiments is 500, for TinyImageNet, PennTreebank and WikiText2 experiments is 800, for WikiText103 experiments is 100. Local epoch for particitants is 1. $\tau=0.5$ like in \cite{chen2020simple}. $\lambda=1$ following \cite{li2021model}. The initial conrol parameter $c_0=0.4$, and final stage parameter $p=0.75$. Table \ref{Experiments details} present the parameter size, data partition, and model architecture for each experiment.
 
Learning rate scheduler for {\ttfamily MSfusion} is CyclicLR scheduler which varies the learning rate between the minimal and maximal thresholds. The learning rate values change in a cycle from more minor to higher and vice versa. The reasons for choosing CyclicLR is the dynamic mechanisms in {\ttfamily MSfusion}. The minimal thresholds is set 0.001, the maximal is set 0.0012, the cycle round is the same with the maximum communication round. Note to ensure equitable comparisons, we maintain uniformity in parameters across all PT-based baselines(HeteroFL and FedRolex), including learning rate, and the number of communication rounds. Other hyperparameters of FedRolex and HeteroFL is set by following \cite{alam2023fedrolex}.

All experiments are conducted using PyTorch version 2.0 on a single machine equipped with two Intel platinum 8378A CPUs, 512GB of memory, and eight NVIDIA A6000 GPUs.

\begin{figure*}[h]
\begin{center}
\centerline{\includegraphics[width=180mm]{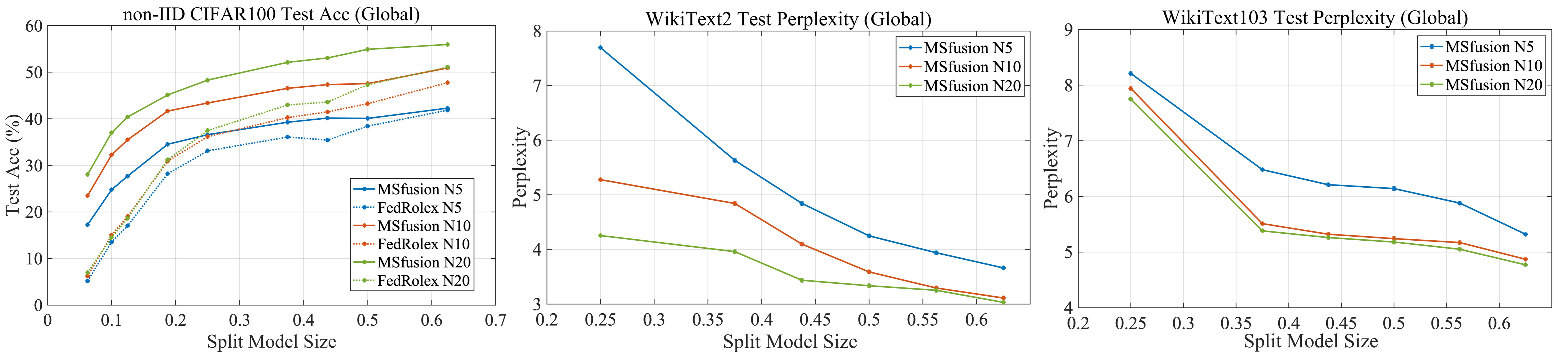}}
% \vspace{-4mm}
\caption{Performance of {\ttfamily MSfusion} for different numbers of participants.}
\vspace{-7mm}
\label{Scalability}
\end{center}
\end{figure*}

\subsection{Performance}
Performance and computational cost comparisons between {\ttfamily MSfusion} and the aforementioned baselines are summarized in Table \ref{Performance_Re}. Performance metrics are evaluated by testing the final global model on the testing dataset. For WikiText103 with 1B global full model, HeteroFL and FedRolex fail to converge. %result is marked as N/A. Performance metrics are evaluated by testing this unsplit global model on the testing dataset. 
To ensure equitable comparisons with centralized methods, we apply fully connected topology for MSfusion in these experiments. For Fed-ET, we also include the computation cost of the server and average it across participants. For FedHM, the computation cost of matrix decomposition is calculated, where we set a fixed and uniform rank ratio $\gamma$ for all clients. And for DepthFL, we set a uniform class depth of $a$ for all clients in CIFAR100 and a uniform class depth of $c$ for all clients in TinyImageNet.

Table \ref{Performance_Re} reveals that, particularly for CIFAR100 datasets, {\ttfamily MSfusion} S outperforms HeteroFL and FedRolex with only $10\%$ of the full model per participant, with less than $20\%$ of the computational cost. Increasing the local split model size directly enhances overall performance, though it comes at the cost of increased computation. {\ttfamily MSfusion} M achieves similar performance with Fed-ET and outperforms all other baselines, in the more complex TinyImageNet dataset. {\ttfamily MSfusion} L outperforms Fed-ET for both image datasets while taking only $20\%$ of the computation. Additionally, it's important to note that {\ttfamily MSfusion} does not require a central server and does not rely on public data, in contrast to KD-based methods. In NLP tasks, {\ttfamily MSfusion} significantly outperforms HeteroFL and FedRolex. The suboptimal performance of HeteroFL and FedRolex in collaborative training larger models is due to massive untrained global model parameters peer round and exacerbated model drift in transformer models, issues unaddressed in their splitting-focused methodologies. 
While these problems can be efficiently tackled by DSS and designed contrastive loss in {\ttfamily MSfusion} further emphasizes the advantage of {\ttfamily MSfusion} in handling larger models.
It is also shown that while DepthFL and FedHM exhibit improved model accuracy over HeteroFL and FedRolex, MSfusion still obtains a better performance with less computation cost.

Performance comparisons with split model size for CIFAR100, WikiText2 and WikiText103 are shown in Figure \ref{Performance Fig}. The perofrmance advantage of {\ttfamily MSfusion} is substantial, especially for smaller split model sizes, due to the more effective DSS scheme. To achieve a target accuracy, much smaller split model size is required for {\ttfamily MSfusion}, resulting in significantly lower computation and communication costs at all participants. Like for CIFAR100, for a target 40\% accuracy performance, FedRolex requires about 45\% split model size which costs 116M FLOPs, while {\ttfamily MSfusion} only requires 12.5\% split model size which only costs 9.83M FLOPs. That is, to train a global model with the same performance,a {\ttfamily MSfusion} participant incurs less than one tens of the computation cost of a FedRolex participant.
% more that 10 times more computation power for each participant to train a same size global model with ruffly the same performance. 

The results of an ablation study on CIFAR100 and WikiText2 with $\mu=18.75\%$ is given in Table~\ref{Ablation studies}. It shows the accuracy and computation comparison for {\ttfamily MSfusion}, {\ttfamily MSfusion} without contrastive objective (w/o Con), {\ttfamily MSfusion} without dynamic overlap (w/o Dyn), and {\ttfamily MSfusion} without both (w/o Con $\&$ Dyn). 
The results showcase {\ttfamily MSfusion}'s superior performance over all baselines. Specifically, the performance gain of {\ttfamily MSfusion} is more prominent for NLP tasks. This is mainly attributed to the relatively larger size of local model split within the transformer architecture, further amplifying the model drift problem.
This study demonstrates that the proposed contrastive objective and dynamic overlap techniques play key roles in ensuring the effectiveness of {\ttfamily MSfusion} training, with a marginal increase in computational cost.
%The contrastive objective cost only few computation power to perform. 
Moreover, we note that {\ttfamily MSfusion} without contrastive objective still outperforms all PT-based methods with much less computation cost.

\begin{table}[h]
% \vspace{-5mm}
\caption{Performance and computational costs of baselines and {\ttfamily MSfusion} variants.}
                \scalebox{0.82}{\begin{tabular}{lcccc}
                        \toprule
                        Methods &  CIFAR100 ACC  & FLOPs & WikiText2 PPL & FLOPs\\
                        \midrule

                        HeteroFL & 12.11 $\pm$ 0.7 & 35.76M & 579.05 $\pm$ 8.4 & 1.08B\\
                        FedRolex & 36.61 $\pm$ 1.0 & 35.76M & 547.32 $\pm$ 4.5 & 1.08B\\
                        MSfusion w/o Con & 40.96 $\pm$ 0.4  & 20.32M & 9.57 $\pm$ 2.1 & 257.2M\\
                        MSfusion w/o Dyn & 41.48 $\pm$ 0.4 & 22.33M & 7.35 $\pm$ 1.5 & 290.1M\\
                        MSfusion w/o Con $\&$  Dyn & 39.77 $\pm$ 0.5 & \textbf{20.32M} & 11.24 $\pm$ 2.3 & \textbf{257.2M}\\
                        MSfusion  & \textbf{44.11 $\pm$ 0.4} & 22.33M & \textbf{5.276 $\pm$ 0.4} & 290.1M\\
                        \bottomrule
                    \end{tabular}
}
% \vspace{-5mm}
			\label{Ablation studies}
\end{table}

\subsection{Scalability}
\label{Scalability_sec}
To assess scalability of {\ttfamily MSfusion}, we conduct experiments on different numbers of participants, where each participant holds $5\%$ of the training dataset.
% conducted experiments where we held the training data size constant for participants, each assigned a fixed 1/20 portion of the training data (while the test data for the global full model remained unchanged). We then evaluated the performance of {\ttfamily MSfusion} and FedRolex with varying numbers of participants, all employing the same split model size for collaborative training of a larger global full model. 
As shown in Figure \ref{Scalability}, {\ttfamily MSfusion} consistently outperforms the SOTA FedRolex for all participant counts. Notably, for small split model sizes, introducing more participants does not help to improve the performance of FedRolex, as it does not properly address the issue of model drift.
% FedRolex faces challenges in ensuring scalability, particularly when dealing with small split model sizes, as it does not address the issue of model drift. 
For {\ttfamily MSfusion}, the model performance improves at all split model sizes as the number of participants increases; to achieve a target level of performance, as participants increases, the required split model size on each participant reduces significantly. 
For instance for WikiText2, to achieve a target PPL around 4, the split model size reduces from 56.26\% with 5 participants to merely 37.5\% with 20 participants. %\textcolor{red}{Use an NLP example}
This makes {\ttfamily MSfusion} a key enabler for resource-constrained machines to contribute to and benefit from collaborative training. 
% Achieving a target level of accuracy requires fewer participants and smaller split model sizes with {\ttfamily MSfusion}.

\begin{figure*}[h]
\begin{center}
\centerline{\includegraphics[width=180mm]{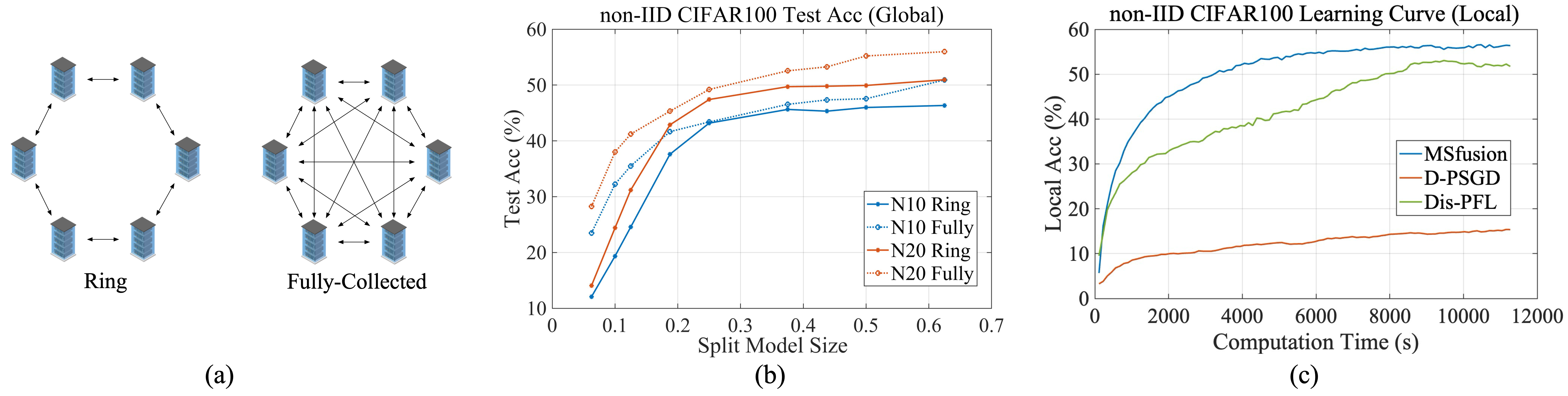}}
\vspace{-3mm}
\caption{(a) Illustrations of ring and fully-connected network topology; (b) Performance of {\ttfamily MSfusion} under ring and fully-connected topology; (c) Performance comparison under ring topology ({\ttfamily MSfusion} $\mu=25\%$ for all participants).}
\vspace{-5mm}
\label{Ring topology}
\end{center}
\end{figure*}

% \vspace{-3mm}
\subsection{Topology}
In scenarios with limited network resources, we may need to utilize a ring topology for participants to communicate. Figure \ref{Ring topology}(a) illustrates the distinction between the ring topology and the fully-connected topology. 
% To apply {\ttfamily MSfusion} in ring topology, 
% the participants with nearest network topologies, the indexing is designed to be proximate, thereby facilitating optimal communication. 
We examine the performance of {\ttfamily MSfusion} under these two network topologies, as detailed in Figure \ref{Ring topology}(b). We find that under the ring topology, {\ttfamily MSfusion} is able to achieve a sizable portion of the full performance achieved used fully-connected topology, and the gap is rather marginal for the most relevant split model sizes of $25\% \sim 45\%$. Also, operating {\ttfamily MSfusion} under the ring topology still attains the desiable scalability with number of participants.
% It is worth noting that the Global Model Combination step in {\ttfamily MSfusion} occurs every 10 communication rounds, primarily for evaluating the performance of the global full model on the testing dataset (this is exclusively for testing purposes and is not required during actual training). 
% Our findings reveal that {\ttfamily MSfusion} effectively facilitates a scalable collaborative training process while maintaining a high level of performance under ring topology.

\begin{table}[h]
% \vspace{-5mm}
    \centering
\caption{Accuracy, communication cost (COMM), and computation cost 
(FLOPs) comparison of CIFAR-100 for ring topology.}
                \scalebox{1}{\begin{tabular}{llcc}
        \toprule
        Methods & Local ACC  & COMM & FLOPs\\
        \midrule
        Dis-PFL & 52.57 $\pm$ 0.3  & 44.8MB  & 700.1M \\
        D-PSGD & 13.38 $\pm$ 0.5 & 89.7MB & 830.3M \\
        MSfusion  & \textbf{56.57 $\pm$ 0.6} & \textbf{2.85MB} & \textbf{22.33M} \\
        % MSfusion(ter)  & \textbf{56.57 $\pm$ 0.6} & \textbf{3.21MB} & \textbf{22.33M}  \\
        \bottomrule
    \end{tabular}
}
% \vspace{-5mm}
\label{Efficiency}
\end{table}

Figure~\ref{Ring topology}(c) shows the performance comparison between {\ttfamily MSfusion}, Dis-PFL, and D-PSGD. As in both Dis-PFL and D-PSGD, each participant retains a personalized model, and accuracy is calculated as the average of local accuracies, we measure the average accuracy of the global model trained using {\ttfamily MSfusion} on local datasets.
% To ensure a fair comparison, {\ttfamily MSfusion} leverages local maintained global full model accuracy over the local dataset, which is then averaged across all participants to determine local accuracy. 
we can see that {\ttfamily MSfusion} converges much faster and outperforms both baselines. 
% The results shown {\ttfamily MSfusion} is able to converge significantly faster while outperforming Dis-PFL. 
% Thanks to its efficient DSS scheme, {\ttfamily MSfusion} achieves competitive performance in less than one-third of the computation time. 
Additionally, the model accuracy, and the average communication and computation costs for CIFAR-100 are compared in Table \ref{Efficiency}. {\ttfamily MSfusion} achieves better performance with less than $5\%$ communication and computation costs, corroborating the effectiveness and efficiency of its overlap averaging mechanism.

\begin{figure}[h]
\label{hetero1}
\begin{center}
\centerline{\includegraphics[width=88mm]{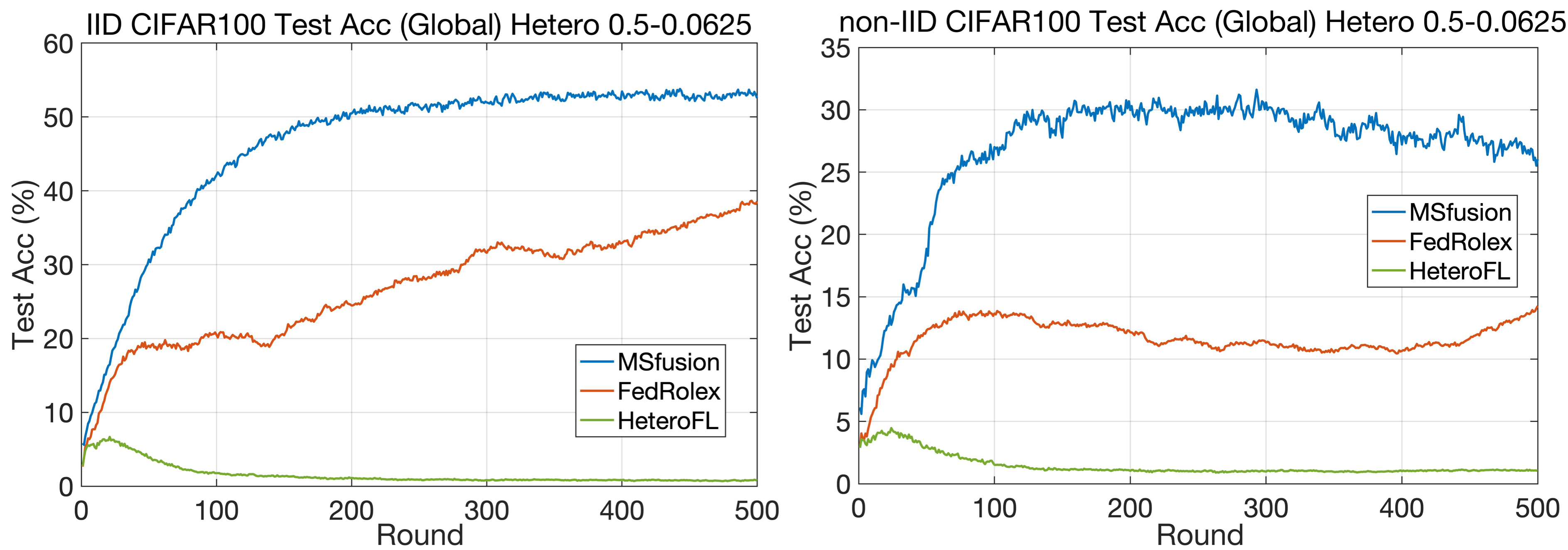}}
% \vspace{-2mm}
\caption{Heterogeneous split model size comparison $(\mu_n\in \{0.5, 0.25, 0.1875, 0.125, 0.0625\})$}
\label{hetero_05_00625}
\end{center}
\vspace{-4mm}
\end{figure}

\begin{figure}[h]
\label{hetero2}
\begin{center}
\centerline{\includegraphics[width=88mm]{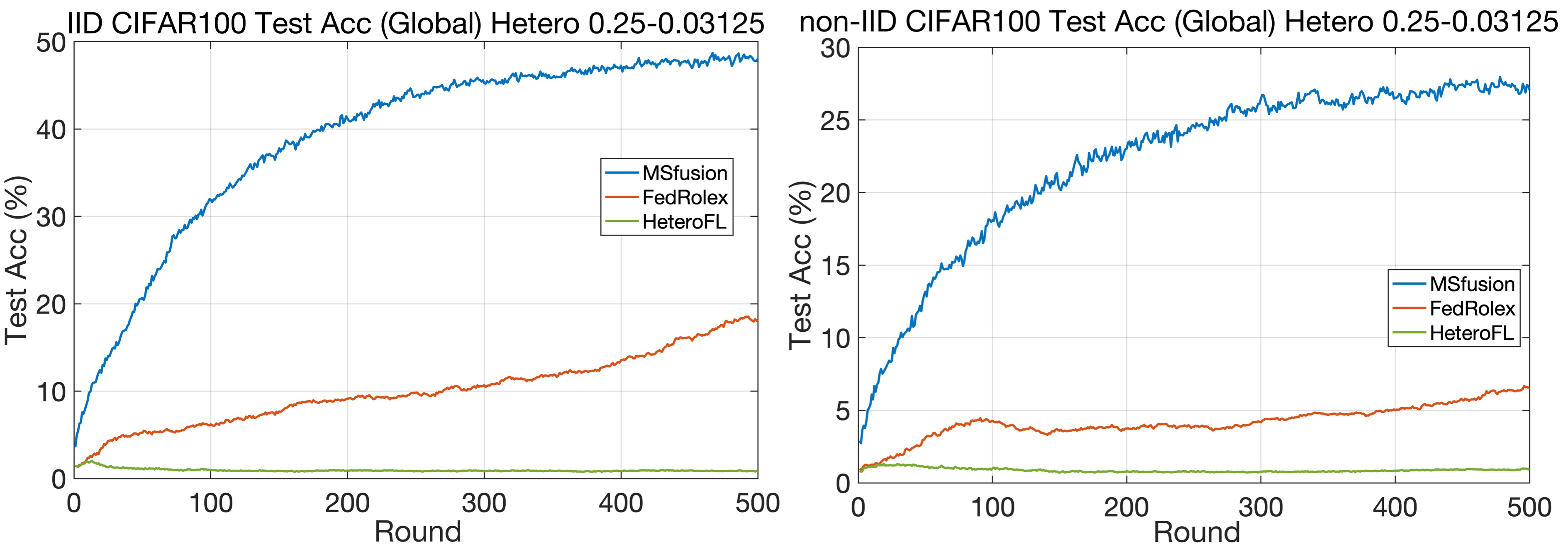}}
% \vspace{-2mm}
\caption{Heterogeneous split model size comparison $(\mu_n\in \{0.25, 0.1875, 0.125, 0.0625, 0.03125\})$}
\label{hetero_025_003125}
\end{center}
\vspace{-4mm}
\end{figure}

\subsection{Heterogeneous split fusion setting}
We also discuss the heterogeneous split fusion setting where particitants' server with different computation power are collaboratively training a large model. Specifically, in this paper, our heterogeneous setting is more constrained with split model size for all participants is less than half of the global full model ($\forall \mu_n \le 0.5$). Figure. \ref{hetero_05_00625} shows the heterogeneous split model size comparison with $(\mu_n\in \{0.5, 0.25, 0.1875, 0.125, 0.0625\})$, Figure. \ref{hetero_025_003125} shows the heterogeneous split model size comparison with $(\mu_n\in \{0.25, 0.1875, 0.125, 0.0625, 0.03125\})$. In these experiments, total of 10 participants are involved in the collaboratively training process, each 2 are assigned with a fixed unique split model size from the list sets. It is clear that the proposed {\ttfamily MSfusion} way outperformed STOA PT-based methods in the more constrained heterogeneous collaboratively learning settings in both IID and non-IID data distributions with much faster converge speed and higher accuracy. FedRolex and outperform HeteroFL with its round-rolling scheme, but the performance of FedRolex is greatly dropped with smaller split model size in Figure. \ref{hetero_025_003125}. While {\ttfamily MSfusion} can still maintain a good performance thanks to its much more efficient DSS scheme. It can also be observed that the model heterogeneous will greatly enhence the model drift between the particitants resulting greatly influence the performance in the non-IID settings. 

% A comprehensive overview of local efficiency, specifically for CIFAR-100, is provided in Table. \ref{Efficiency}. 

% Notably, owing to the application of dynamic overlap in {\ttfamily MSfusion}, there exists slight discrepancy in communication costs between each round, a averaged value is provided.
\begin{figure}[h]
% \vspace{-3mm}
\begin{center}
\centerline{\includegraphics[width=88mm]{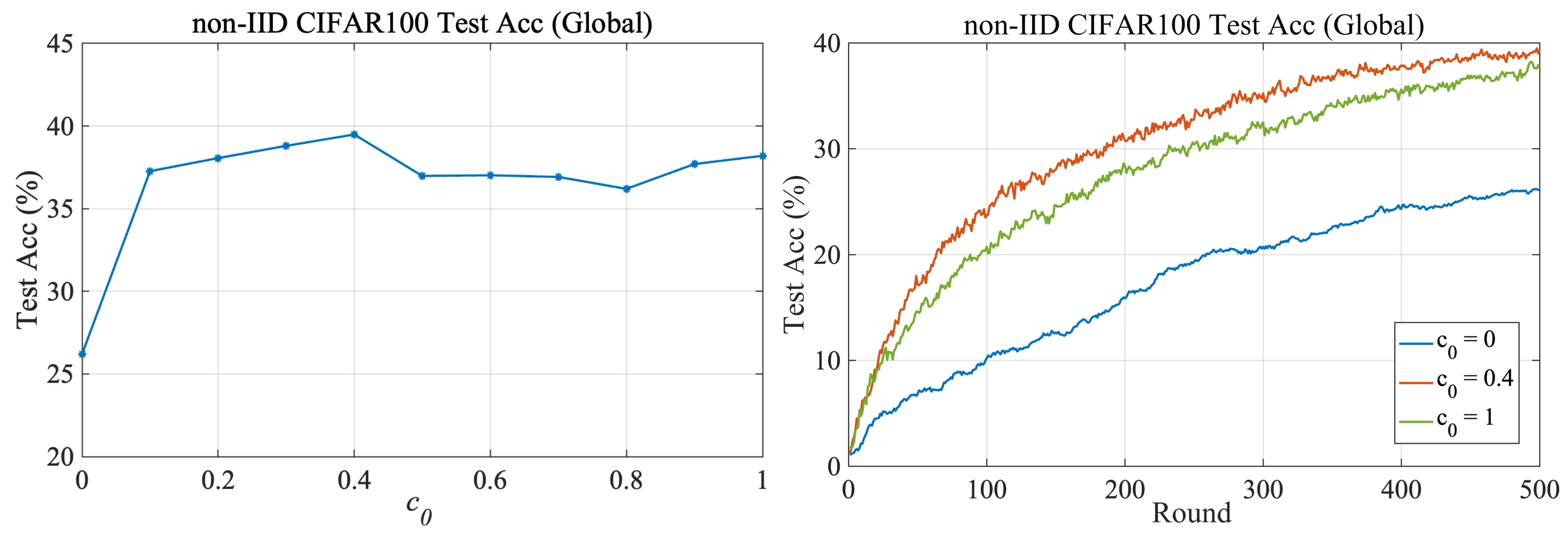}}
% \vspace{-4mm}
\caption{Effect of initial overlapping control parameter $c_0$.}
\label{Effect of c0}
\end{center}
\vspace{-4mm}
\end{figure}

\begin{figure}[h]
\begin{center}
\centerline{\includegraphics[width=88mm]{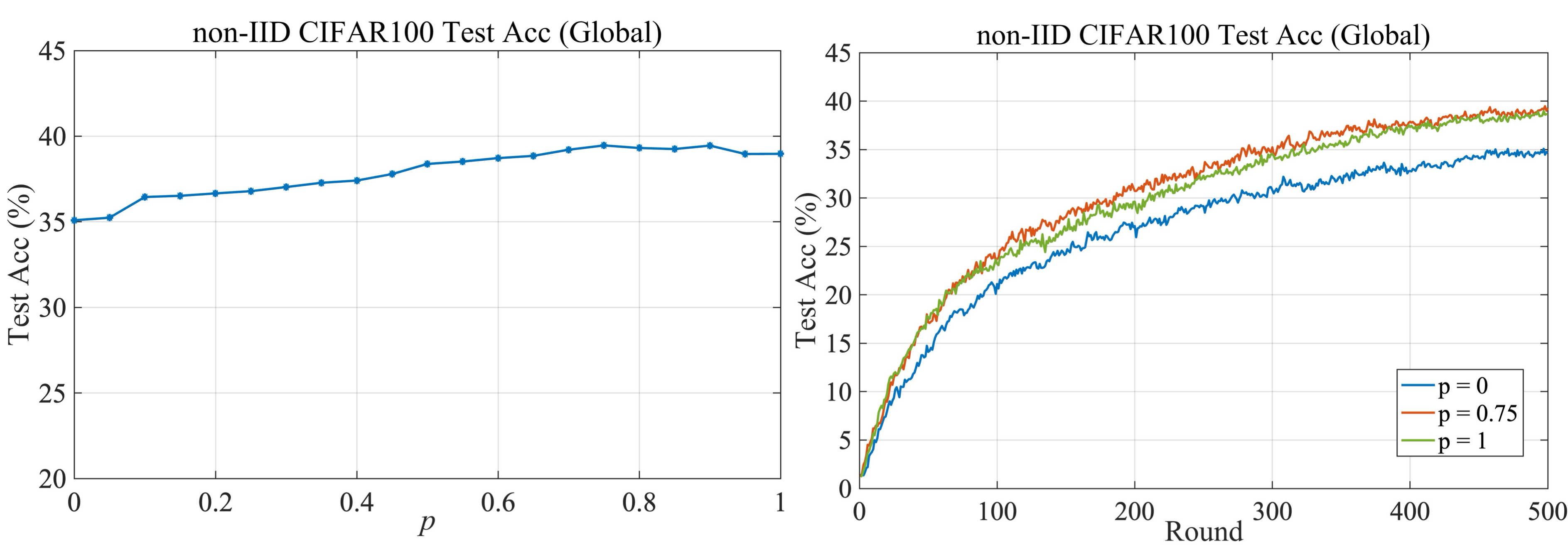}}
\caption{Effect of overlapping final stage parameter $p$.}
\label{Effect of p}
\end{center}
\vspace{-4mm}
\end{figure}

    \begin{table*}[h]
% \vspace{-5mm}
    \centering
\caption{MSfusion performance of full model and local split model.}
                \scalebox{0.85}{\begin{tabular}{lcccccc}
                \toprule
                                      & \multicolumn{3}{c}{non-iid~CIFAR100}           & \multicolumn{3}{c}{WikiText2}                \\
                                      \midrule
Method                                & ACC               & Split ACC       & FLOPs  & Perplexity      & Split Perplexity & Size    \\
\midrule
MSfusion $\mu$ = 100\%   & 52.61~$\pm$~0.7   & 52.61~$\pm$~0.7             & 572.2M & 3.26~$\pm$~0.2  & 3.26~$\pm$~0.2              & 139.01M \\
MSfusion $\mu$ =~18.75\% & 44.11 $\pm$~0.4   & 39.54~$\pm$~0.4 & 22.33M & 44.33~$\pm$~3.6 & 68.73 $\pm$ 5.4  & 30.33M  \\
MSfusion $\mu$ = 50\%    & 47.21~$\pm$~0.5   & 41.28~$\pm$~0.6 & 151.7M & 3.59~$\pm$~0.2  & 5.69~$\pm$~0.5   & 69.23M  \\
\bottomrule
\end{tabular}
}
\vspace{-5mm}
\label{Full compare}
\end{table*}

\subsection{Ablation study}
An ablation study toward the effect of initial overlapping control parameter $c_0$ is given in Figure \ref{Effect of c0}. 
% We can see that making $c_0$ too small may cause an insufficient coverage of the full model at the beginning of training, using large $c_0$ may suffer more severely from model drift. Here the optimal test accuracy is achieved at some intermediate value of $c_0 = 0.4$. 
Specifically, it demonstrates that a $c_0=1$ setting leads to slightly faster convergence than $c_0=0.4$ during the initial stages of training, both outperforming the $c=0$ scenario where the global full model shows insufficient training initially. This evidence supports the insufficient convergence at the beginning of training. Furthermore, thanks to a larger overlap in parameters between local models and hence reduced model drift, $c_0=0.4$ starts to outperform $c_0=1$ after the initial stages of training.

We also provide ablation study on the effect of parameter $p$ that controls the final overlap rate (i.e., the overlap rate $c = c_0 (1-\frac{r}{R}p)$ at round $r$), reflected in Figure \ref{Effect of p}. Specifically, we observed that a $p$ value of 0, which implies a constant $c$, leads to performance degradation due to client drift. Conversely, increasing $p$ narrows the inter-participant gap as the training progresses, enhancing the model's ability to leverage data from a broader range of participants. However, excessively high $p$ values can hinder convergence speed, as evidenced by the learning curve where $p=0.75$ achieves faster convergence over $p=1$. Thus, we identified $p=0.75$ as the optimal parameter for a faster convergence to a higher accuracy. 
% It can be shown that there is a huge performance difference between without inter-participant gap ($c_0=0$) and with inter-participant gap. And the performance gradually increase until a optimal $c_0^*$, this shown there is a trade-off between overlapping rate and global full model coverage each round with all participants.

\begin{table}[h]
% \vspace{-5mm}
    \centering
\caption{MSfusion performance with difference sequence lengths (smaller perplexity is better), split model size $\mu = 25\%$.}
                \scalebox{0.80}{\begin{tabular}{llll}
                \toprule
                & PennTreebank                 & WikiText2                   & WikiText103                  \\
                \midrule
Sequence Length & Perplexity                   & Perplexity                  & Perplexity                   \\
\midrule
64              & 8.02~$\pm$~0.5  & 5.28~$\pm$~0.4 & 7.94~$\pm$~0.6  \\
256             & 8.41 $\pm$~0.4  & 5.75 $\pm$~0.4 & 8.14~$\pm$~0.6  \\
512             & 10.11 $\pm$~0.6 & 6.31~$\pm$~0.5 & 9.47 $\pm$~0.7  \\
\bottomrule
\end{tabular}
}
%\vspace{-5mm}
\label{Sequence Length}
\end{table}

\subsection{Additional experiments}

\textbf{Performance of full model and local split model.} We also evaluated the performance of model without splitting ($\mu = 100\%$) and the local split model performance on the global test dataset. Results are shown in Table~\ref{Full compare}. The ACC and Perplexity is the performance of combined global full model, and Split ACC and Split Perplexity is the local split model performance averaged over all participants on the global test dataset. 

We observe that MSfusion with $\mu=50\%$ achieves a comparable performance with $\mu = 100\%$, with much less computation cost. This tradeoff is acceptable or even desirable in scenarios where the training is limited by the computation, storage and communicaiton resources. Besides, for $\mu < 100\%$, there is a sizable performance loss on the local split model. This is due to 1) larger number of parameters in the full model; and 2) training over other participants' datasets.

\textbf{Different sequence length.} We adopted a sequence length of 64 to make a fair comparison with prior works on federated learning with model splitting ~\cite{diao2021heterofl, alam2023fedrolex, kim2022depthfl}, which used a sequence length of 64 in their experiments. Another reason to use a smaller sequence length is to reduce computation complexities, since this work focuses on resource-constrained participants. However, we also note that more commonly used sequence lengths should be considered, and we provided experiments on sequence lengths of 256 and 512. Results are reported in Table~\ref{Sequence Length}, it indicates that MSfusion maintains its effectiveness across a broader range of sequence lengths, achieving comparable performance as the sequence length increases. This result underscores the robustness of MSfusion in accommodating various NLP task requirements.

% We have also evaluated {\ttfamily MSfusion} for heterogeneous split model sizes across participants, and experimentally shown that {\ttfamily MSfusion} achieves good performance and outperforms baselines for both iid and non-iid settings. We present experiment details in Appendix~
% \ref{sec:Heterogeneous} and results in Figure~\ref{hetero_05_00625} and \ref{hetero_025_003125}.

% ). As shown in Figure~\ref{hetero_05_00625} and \ref{hetero_025_003125} {\ttfamily MSfusion} achieves good performance and outperforms baselines for both iid and non-iid settings.
% We also report a discussion toward heterogeneous split fusion setting in Appendix \ref{sec:Heterogeneous}. Showing that {\ttfamily MSfusion} can still maintain a good performance in this setting.

% These findings underscore that {\ttfamily MSfusion} entails only 1/20 of the communication cost and 1/20 of the computation cost, a testament to its efficient overlap averaging methodology.

% \vspace{-2mm}
\section{Conclusion and discussion}
\label{others}
We introduce {\ttfamily MSfusion} to address the challenge in collaboratively training large models on resource-constrained machines. Through an ensemble of novel techniques like DSS, dynamic model split and aggregation, and contrastive loss design, {\ttfamily MSfusion} achieves superior performance with significantly reduced complexity. %Its adaptive overlap strategy coupled with a tailored contrastive objective further sets it apart by ensuring convergence speed and mitigating model drift, respectively. 
Our empirical results underscore the potential of {\ttfamily MSfusion} to collaboratively training large models for CV and NLP tasks, offering a promising solution for organizations striving to maximize performance while judiciously managing resources. 

While the primary focus of MSfusion is on optimizing the performance of training larger models in a resource-constrained, decentralized environment, we also recognize the importance of privacy concerns, and have designed MSfusion to be readily compatible with well-established privacy-preserving mechanisms like Differential Privacy (DP) and Secure Aggregation~\cite{bonawitz2017practical,so2022lightsecagg}, further protecting participants' data privacy.

% \section{Extension}
% In this paper, the primary focus with MSfusion is on optimizing the training of larger models in a resource-constrained, decentralized environment. Howerver, we also recognize the importance of privacy concerns and have designed MSfusion to be readily compatible with well-established defense mechanisms like Differential Privacy (DP) and Secure Aggregation [4], further protecting participants' data privacy.
    
% While the initial implementation of MSfusion demonstrates potential advantages in privacy and security compared to traditional centralized methods, we agree that a comprehensive study of its vulnerability for specific attacks and countermeasures warrant further research. As such, we aim to dedicate future work to exploring and enhancing the privacy and security aspects of MSfusion, ensuring that it not only facilitates efficient training of large models, but also adheres to a higher standard of participant data protection.

\bibliography{Ref}

\begin{thebibliography}{10}
\providecommand{\url}[1]{#1}
\csname url@samestyle\endcsname
\providecommand{\newblock}{\relax}
\providecommand{\bibinfo}[2]{#2}
\providecommand{\BIBentrySTDinterwordspacing}{\spaceskip=0pt\relax}
\providecommand{\BIBentryALTinterwordstretchfactor}{4}
\providecommand{\BIBentryALTinterwordspacing}{\spaceskip=\fontdimen2\font plus
\BIBentryALTinterwordstretchfactor\fontdimen3\font minus \fontdimen4\font\relax}
\providecommand{\BIBforeignlanguage}[2]{{%
\expandafter\ifx\csname l@#1\endcsname\relax
\typeout{** WARNING: IEEEtran.bst: No hyphenation pattern has been}%
\typeout{** loaded for the language `#1'. Using the pattern for}%
\typeout{** the default language instead.}%
\else
\language=\csname l@#1\endcsname
\fi
#2}}
\providecommand{\BIBdecl}{\relax}
\BIBdecl

\bibitem{brown2020language}
T.~Brown, B.~Mann, N.~Ryder, M.~Subbiah, J.~D. Kaplan, P.~Dhariwal, A.~Neelakantan, P.~Shyam, G.~Sastry, A.~Askell \emph{et~al.}, ``Language models are few-shot learners,'' \emph{Advances in neural information processing systems}, vol.~33, pp. 1877--1901, 2020.

\bibitem{floridi2020gpt}
L.~Floridi and M.~Chiriatti, ``Gpt-3: Its nature, scope, limits, and consequences,'' \emph{Minds and Machines}, vol.~30, pp. 681--694, 2020.

\bibitem{zhang2022opt}
S.~Zhang, S.~Roller, N.~Goyal, M.~Artetxe, M.~Chen, S.~Chen, C.~Dewan, M.~Diab, X.~Li, X.~V. Lin \emph{et~al.}, ``Opt: Open pre-trained transformer language models,'' \emph{arXiv preprint arXiv:2205.01068}, 2022.

\bibitem{chowdhery2023palm}
A.~Chowdhery, S.~Narang, J.~Devlin, M.~Bosma, G.~Mishra, A.~Roberts, P.~Barham, H.~W. Chung, C.~Sutton, S.~Gehrmann \emph{et~al.}, ``Palm: Scaling language modeling with pathways,'' \emph{Journal of Machine Learning Research}, vol.~24, no. 240, pp. 1--113, 2023.

\bibitem{achiam2023gpt}
J.~Achiam, S.~Adler, S.~Agarwal, L.~Ahmad, I.~Akkaya, F.~L. Aleman, D.~Almeida, J.~Altenschmidt, S.~Altman, S.~Anadkat \emph{et~al.}, ``Gpt-4 technical report,'' \emph{arXiv preprint arXiv:2303.08774}, 2023.

\bibitem{radford2018improving}
A.~Radford, K.~Narasimhan, T.~Salimans, I.~Sutskever \emph{et~al.}, ``Improving language understanding by generative pre-training,'' 2018.

\bibitem{devlin2018bert}
J.~Devlin, M.-W. Chang, K.~Lee, and K.~Toutanova, ``Bert: Pre-training of deep bidirectional transformers for language understanding,'' \emph{arXiv preprint arXiv:1810.04805}, 2018.

\bibitem{raffel2020exploring}
C.~Raffel, N.~Shazeer, A.~Roberts, K.~Lee, S.~Narang, M.~Matena, Y.~Zhou, W.~Li, and P.~J. Liu, ``Exploring the limits of transfer learning with a unified text-to-text transformer,'' \emph{Journal of machine learning research}, vol.~21, no. 140, pp. 1--67, 2020.

\bibitem{bommasani2021opportunities}
R.~Bommasani, D.~A. Hudson, E.~Adeli, R.~Altman, S.~Arora, S.~von Arx, M.~S. Bernstein, J.~Bohg, A.~Bosselut, E.~Brunskill \emph{et~al.}, ``On the opportunities and risks of foundation models,'' \emph{arXiv preprint arXiv:2108.07258}, 2021.

\bibitem{zhao2023survey}
W.~X. Zhao, K.~Zhou, J.~Li, T.~Tang, X.~Wang, Y.~Hou, Y.~Min, B.~Zhang, J.~Zhang, Z.~Dong \emph{et~al.}, ``A survey of large language models,'' \emph{arXiv preprint arXiv:2303.18223}, 2023.

\bibitem{mcmahan2017communication}
B.~McMahan, E.~Moore, D.~Ramage, S.~Hampson, and B.~A. y~Arcas, ``Communication-efficient learning of deep networks from decentralized data,'' in \emph{Artificial intelligence and statistics}.\hskip 1em plus 0.5em minus 0.4em\relax PMLR, 2017, pp. 1273--1282.

\bibitem{dey2023cerebrasgpt}
N.~Dey, G.~Gosal, Zhiming, Chen, H.~Khachane, W.~Marshall, R.~Pathria, M.~Tom, and J.~Hestness, ``Cerebras-gpt: Open compute-optimal language models trained on the cerebras wafer-scale cluster,'' 2023.

\bibitem{wang2022accelerating}
L.~Wang, Y.~Xu, H.~Xu, M.~Chen, and L.~Huang, ``Accelerating decentralized federated learning in heterogeneous edge computing,'' \emph{IEEE Transactions on Mobile Computing}, 2022.

\bibitem{yuan2024decentralized}
L.~Yuan, Z.~Wang, L.~Sun, S.~Y. Philip, and C.~G. Brinton, ``Decentralized federated learning: A survey and perspective,'' \emph{IEEE Internet of Things Journal}, 2024.

\bibitem{chen2024enhancing}
S.~Chen, Y.~Xu, H.~Xu, Z.~Ma, and Z.~Wang, ``Enhancing decentralized and personalized federated learning with topology construction,'' \emph{IEEE Transactions on Mobile Computing}, no.~01, pp. 1--16, 2024.

\bibitem{sun2022decentralized}
T.~Sun, D.~Li, and B.~Wang, ``Decentralized federated averaging,'' \emph{IEEE Transactions on Pattern Analysis and Machine Intelligence}, vol.~45, no.~4, pp. 4289--4301, 2022.

\bibitem{lian2017can}
X.~Lian, C.~Zhang, H.~Zhang, C.-J. Hsieh, W.~Zhang, and J.~Liu, ``Can decentralized algorithms outperform centralized algorithms? a case study for decentralized parallel stochastic gradient descent,'' \emph{Advances in neural information processing systems}, vol.~30, 2017.

\bibitem{roy2019braintorrent}
A.~G. Roy, S.~Siddiqui, S.~P{\"o}lsterl, N.~Navab, and C.~Wachinger, ``Braintorrent: A peer-to-peer environment for decentralized federated learning,'' \emph{arXiv preprint arXiv:1905.06731}, 2019.

\bibitem{dai2022dispfl}
R.~Dai, L.~Shen, F.~He, X.~Tian, and D.~Tao, ``Dispfl: Towards communication-efficient personalized federated learning via decentralized sparse training,'' \emph{arXiv preprint arXiv:2206.00187}, 2022.

\bibitem{gou2021knowledge}
J.~Gou, B.~Yu, S.~J. Maybank, and D.~Tao, ``Knowledge distillation: A survey,'' \emph{International Journal of Computer Vision}, vol. 129, no.~6, pp. 1789--1819, 2021.

\bibitem{cho2019efficacy}
J.~H. Cho and B.~Hariharan, ``On the efficacy of knowledge distillation,'' in \emph{Proceedings of the IEEE/CVF international conference on computer vision}, 2019, pp. 4794--4802.

\bibitem{mirzadeh2020improved}
S.~I. Mirzadeh, M.~Farajtabar, A.~Li, N.~Levine, A.~Matsukawa, and H.~Ghasemzadeh, ``Improved knowledge distillation via teacher assistant,'' in \emph{Proceedings of the AAAI conference on artificial intelligence}, vol.~34, no.~04, 2020, pp. 5191--5198.

\bibitem{zhou2023mec}
X.~Zhou, Y.~Tian, and X.~Wang, ``Mec-da: Memory-efficient collaborative domain adaptation for mobile edge devices,'' \emph{IEEE Transactions on Mobile Computing}, 2023.

\bibitem{cho2022heterogeneous}
Y.~J. Cho, A.~Manoel, G.~Joshi, R.~Sim, and D.~Dimitriadis, ``Heterogeneous ensemble knowledge transfer for training large models in federated learning,'' 2022.

\bibitem{lin2020ensemble}
T.~Lin, L.~Kong, S.~U. Stich, and M.~Jaggi, ``Ensemble distillation for robust model fusion in federated learning,'' \emph{Advances in Neural Information Processing Systems}, vol.~33, pp. 2351--2363, 2020.

\bibitem{hong2022efficient}
J.~Hong, H.~Wang, Z.~Wang, and J.~Zhou, ``Efficient split-mix federated learning for on-demand and in-situ customization,'' \emph{arXiv preprint arXiv:2203.09747}, 2022.

\bibitem{alam2023fedrolex}
S.~Alam, L.~Liu, M.~Yan, and M.~Zhang, ``Fedrolex: Model-heterogeneous federated learning with rolling sub-model extraction,'' 2023.

\bibitem{diao2021heterofl}
E.~Diao, J.~Ding, and V.~Tarokh, ``Heterofl: Computation and communication efficient federated learning for heterogeneous clients,'' 2021.

\bibitem{shulgin2023towards}
E.~Shulgin and P.~Richt{\'a}rik, ``Towards a better theoretical understanding of independent subnetwork training,'' \emph{arXiv preprint arXiv:2306.16484}, 2023.

\bibitem{vepakomma2018split}
P.~Vepakomma, O.~Gupta, T.~Swedish, and R.~Raskar, ``Split learning for health: Distributed deep learning without sharing raw patient data,'' \emph{arXiv preprint arXiv:1812.00564}, 2018.

\bibitem{singh2019detailed}
A.~Singh, P.~Vepakomma, O.~Gupta, and R.~Raskar, ``Detailed comparison of communication efficiency of split learning and federated learning,'' \emph{arXiv preprint arXiv:1909.09145}, 2019.

\bibitem{gao2020end}
Y.~Gao, M.~Kim, S.~Abuadbba, Y.~Kim, C.~Thapa, K.~Kim, S.~A. Camtepe, H.~Kim, and S.~Nepal, ``End-to-end evaluation of federated learning and split learning for internet of things,'' \emph{arXiv preprint arXiv:2003.13376}, 2020.

\bibitem{thapa2022splitfed}
C.~Thapa, P.~C.~M. Arachchige, S.~Camtepe, and L.~Sun, ``Splitfed: When federated learning meets split learning,'' in \emph{Proceedings of the AAAI Conference on Artificial Intelligence}, vol.~36, no.~8, 2022, pp. 8485--8493.

\bibitem{chen2022fedobd}
Y.~Chen, Z.~Chen, P.~Wu, and H.~Yu, ``Fedobd: Opportunistic block dropout for efficiently training large-scale neural networks through federated learning,'' \emph{arXiv preprint arXiv:2208.05174}, 2022.

\bibitem{zhang2023gpt}
T.~Zhang, T.~Feng, S.~Alam, M.~Zhang, S.~S. Narayanan, and S.~Avestimehr, ``Gpt-fl: Generative pre-trained model-assisted federated learning,'' \emph{arXiv preprint arXiv:2306.02210}, 2023.

\bibitem{li2021ditto}
T.~Li, S.~Hu, A.~Beirami, and V.~Smith, ``Ditto: Fair and robust federated learning through personalization,'' in \emph{International conference on machine learning}.\hskip 1em plus 0.5em minus 0.4em\relax PMLR, 2021, pp. 6357--6368.

\bibitem{fallah2020personalized}
A.~Fallah, A.~Mokhtari, and A.~Ozdaglar, ``Personalized federated learning: A meta-learning approach,'' \emph{arXiv preprint arXiv:2002.07948}, 2020.

\bibitem{collins2021exploiting}
L.~Collins, H.~Hassani, A.~Mokhtari, and S.~Shakkottai, ``Exploiting shared representations for personalized federated learning,'' in \emph{International conference on machine learning}.\hskip 1em plus 0.5em minus 0.4em\relax PMLR, 2021, pp. 2089--2099.

\bibitem{beznosikov2020biased}
A.~Beznosikov, S.~Horv{\'a}th, P.~Richt{\'a}rik, and M.~Safaryan, ``On biased compression for distributed learning,'' \emph{arXiv preprint arXiv:2002.12410}, 2020.

\bibitem{safaryan2021smoothness}
M.~Safaryan, F.~Hanzely, and P.~Richt{\'a}rik, ``Smoothness matrices beat smoothness constants: Better communication compression techniques for distributed optimization,'' \emph{Advances in Neural Information Processing Systems}, vol.~34, pp. 25\,688--25\,702, 2021.

\bibitem{wang2022theoretically}
B.~Wang, M.~Safaryan, and P.~Richt{\'a}rik, ``Theoretically better and numerically faster distributed optimization with smoothness-aware quantization techniques,'' \emph{Advances in Neural Information Processing Systems}, vol.~35, pp. 9841--9852, 2022.

\bibitem{chen2020simple}
T.~Chen, S.~Kornblith, M.~Norouzi, and G.~Hinton, ``A simple framework for contrastive learning of visual representations,'' in \emph{International conference on machine learning}.\hskip 1em plus 0.5em minus 0.4em\relax PMLR, 2020, pp. 1597--1607.

\bibitem{he2020momentum}
K.~He, H.~Fan, Y.~Wu, S.~Xie, and R.~Girshick, ``Momentum contrast for unsupervised visual representation learning,'' in \emph{Proceedings of the IEEE/CVF conference on computer vision and pattern recognition}, 2020, pp. 9729--9738.

\bibitem{khosla2020supervised}
P.~Khosla, P.~Teterwak, C.~Wang, A.~Sarna, Y.~Tian, P.~Isola, A.~Maschinot, C.~Liu, and D.~Krishnan, ``Supervised contrastive learning,'' \emph{Advances in neural information processing systems}, vol.~33, pp. 18\,661--18\,673, 2020.

\bibitem{li2021model}
Q.~Li, B.~He, and D.~Song, ``Model-contrastive federated learning,'' in \emph{Proceedings of the IEEE/CVF conference on computer vision and pattern recognition}, 2021, pp. 10\,713--10\,722.

\bibitem{yu2023multimodal}
Q.~Yu, Y.~Liu, Y.~Wang, K.~Xu, and J.~Liu, ``Multimodal federated learning via contrastive representation ensemble,'' \emph{arXiv preprint arXiv:2302.08888}, 2023.

\bibitem{wang2021understanding}
F.~Wang and H.~Liu, ``Understanding the behaviour of contrastive loss,'' in \emph{Proceedings of the IEEE/CVF conference on computer vision and pattern recognition}, 2021, pp. 2495--2504.

\bibitem{krizhevsky2009learning}
A.~Krizhevsky, G.~Hinton \emph{et~al.}, ``Learning multiple layers of features from tiny images,'' 2009.

\bibitem{le2015tiny}
Y.~Le and X.~Yang, ``Tiny imagenet visual recognition challenge,'' \emph{CS 231N}, vol.~7, no.~7, p.~3, 2015.

\bibitem{kim2022depthfl}
M.~Kim, S.~Yu, S.~Kim, and S.-M. Moon, ``Depthfl: Depthwise federated learning for heterogeneous clients,'' in \emph{The Eleventh International Conference on Learning Representations}, 2022.

\bibitem{marcus-etal-1993-building}
\BIBentryALTinterwordspacing
M.~P. Marcus, B.~Santorini, and M.~A. Marcinkiewicz, ``Building a large annotated corpus of {E}nglish: The {P}enn {T}reebank,'' \emph{Computational Linguistics}, vol.~19, no.~2, pp. 313--330, 1993. [Online]. Available: \url{https://aclanthology.org/J93-2004}
\BIBentrySTDinterwordspacing

\bibitem{yao2022fedhm}
D.~Yao, W.~Pan, M.~J. O'Neill, Y.~Dai, Y.~Wan, H.~Jin, and L.~Sun, ``Fedhm: Efficient federated learning for heterogeneous models via low-rank factorization,'' 2022.

\bibitem{bonawitz2017practical}
K.~Bonawitz, V.~Ivanov, B.~Kreuter, A.~Marcedone, H.~B. McMahan, S.~Patel, D.~Ramage, A.~Segal, and K.~Seth, ``Practical secure aggregation for privacy-preserving machine learning,'' in \emph{proceedings of the 2017 ACM SIGSAC Conference on Computer and Communications Security}, 2017, pp. 1175--1191.

\bibitem{so2022lightsecagg}
J.~So, C.~He, C.-S. Yang, S.~Li, Q.~Yu, R.~E~Ali, B.~Guler, and S.~Avestimehr, ``Lightsecagg: a lightweight and versatile design for secure aggregation in federated learning,'' \emph{Proceedings of Machine Learning and Systems}, vol.~4, pp. 694--720, 2022.

\end{thebibliography}
\bibliographystyle{IEEEtran}

\newpage

 

% \end{IEEEbiography}



\vfill

\end{document}